\documentclass[letterpaper, 10 pt, conference]{ieeeconf}  

\IEEEoverridecommandlockouts                              

\usepackage{amsmath,amsfonts}
\usepackage{algorithmic}
\usepackage{algorithm}
\usepackage{array}
\usepackage[caption=false,font=normalsize,labelfont=sf,textfont=sf]{subfig}
\usepackage{textcomp}
\usepackage{stfloats}
\usepackage{url}
\usepackage{verbatim}
\usepackage{graphicx}
\usepackage{booktabs} 
\usepackage{makecell}
\usepackage{cite}
\usepackage[dvipsnames]{xcolor}
\usepackage{lipsum}

\makeatletter
\let\NAT@parse\undefined
\makeatother
\usepackage{multirow}
\usepackage{hyperref}
\usepackage{tabularx}
\usepackage{listings}
\usepackage{placeins}

\newcolumntype{Y}{>{\centering\arraybackslash}X}

\newlength{\figurecaptionskip}
\newlength{\tablecaptionskip}
\makeatletter
\long\def\@makecaption#1#2{%
\ifx\@captype\@IEEEtablestring%
\vskip\tablecaptionskip%
{\footnotesize #1.~~ #2\par}%
\vskip\tablecaptionskip%
\else
\vskip\figurecaptionskip%
\setbox\@tempboxa\hbox{\footnotesize #1.~~ #2}%
\ifdim \wd\@tempboxa >\hsize%
\setbox\@tempboxa\hbox{\footnotesize #1.~~ }%
\parbox[t]{\hsize}{\footnotesize \noindent\unhbox\@tempboxa#2}%
\else%
\ifcenterfigcaptions \hbox to\hsize{\footnotesize\hfil\box\@tempboxa\hfil}%
\else \hbox to\hsize{\footnotesize\box\@tempboxa\hfil}%
\fi\fi\fi}
\makeatother

\hypersetup{
    colorlinks,
    linkcolor={red!50!black},
    citecolor={red!50!black},
    urlcolor={blue!80!black}
}

\newif\ifanonymous
\anonymousfalse  

\newif\ifappendixincluded
\appendixincludedtrue  

\newcommand{\appendixnote}[1]{\ifappendixincluded#1\fi}

\title{\LARGE \bf
On the Generalization Capabilities, Design Choices and Limitations of Keypoint Imitation Learning}

\ifanonymous
\author{Anonymous$^{*}$ - Double Blind Submission
\thanks{*Affiliations anonymized for double-blind submission}%
}
\else
\author{Thomas Lips$^{*,1}$, Marco Moletta$^{*,2}$, Michael C. Welle$^{2,3}$, Danica Kragic$^{2}$ and Francis wyffels$^{1}$
\thanks{*Shared first authors. thomas.lips@UGent.be, moletta@kth.se}
\thanks{$^{1}$ AI and Robotics Lab, IDLAB-AIRO, Ghent University-imec}%
\thanks{$^{2}$ Robotics, Perception and Learning Lab, (RPL), EECS, KTH Royal Institute of Technology.}%
\thanks{$^{3}$ INCAR Robotics AB, Stockholm, Sweden. }%
}
\fi

\begin{document}

\maketitle
\thispagestyle{empty}
\pagestyle{empty}

\begin{abstract}
RGB-based imitation learning requires many demonstrations to generalize to unseen objects or scenes, motivating research into intermediate representations to improve generalization for robotic manipulation.
Visual foundation models enable one-shot extraction of keypoints to provide such representation. However, it remains unclear how to integrate them into imitation learning optimally and when they outperform alternative representations.
We combine approaches from previous works on keypoint imitation learning (KIL) and investigate several design choices to provide practical guidelines. Using over 2000 real-world rollouts, we also assess the generalization capabilities of KIL to unseen objects and scene variations. KIL achieves a 75\% overall success rate across five tasks, significantly outperforming the RGB baseline (47\%) and performing on par with S$^2$-diffusion (73\%). Finally, we explore the limitations of the foundation models used for keypoint extraction and extend KIL to tasks with multiple object instances.
Our results confirm KIL as a data-efficient approach for robot learning, though it does not outperform alternative representations and inherits limitations of the foundation models used for keypoint extraction. 
\end{abstract}

\section{Introduction}
\label{sec:introduction}



Robotic manipulation aims to enable robots to reliably execute physical tasks in unstructured environments, requiring generalization to different poses, unseen objects and scene variations. Imitation learning is a promising approach to achieve this~\cite{chi2025diffusion_policy,barreiros2025LBM-1}, but RGB-based policies still require large amounts of data to handle the aforementioned task variations\cite{barreiros2025LBM-1,lin2025imitation_scaling_laws}. This motivates research into intermediate representations for improved generalization~\cite{ze20243d,yang2025S2,zhu2023viola,zhu2023learning,wang2025skil,di2024KAT,Levy2025P3PO}.

Keypoints are an attractive option for such representation: they are compact, encode fine-grained semantics, and can handle intra-category object variations~\cite{manuelli2019kpam,adriaens2025spill}. Previously, obtaining keypoints required training task-specific detectors, limiting their scalability~\cite{adriaens2025spill,manuelli2019kpam,vecerik2021s3k}. Advances in vision foundation models~\cite{simeoni2025dinov3,heinrich2025radiov2.5,rombach2022stable-diffusion} now enable extracting keypoints using a single annotation and without task-specific training~\cite{amir2021ViTFeatures,tang2023dift}. Several works have leveraged this for keypoint-conditioned imitation learning~\cite{di2024KAT,wang2025skil,Levy2025P3PO,haldar2025pointpolicy}, demonstrating improved performance over raw RGB representations. However, many open questions remain, including how to best extract these keypoints, how to encode them for imitation learning and in which settings this approach outperforms other representations.

This paper consolidates and builds on the methods proposed in previous works on keypoint-conditioned imitation learning (KIL)~\cite{di2024KAT,wang2025skil,Levy2025P3PO,haldar2025pointpolicy} and answers key questions about this approach. To do so, we evaluate the generalization capabilities, describe failure modes, investigate several design choices and expand the scope of KIL to tasks with multiple instances of the same object category.
Our pipeline for keypoint imitation learning consists of three parts. We start by manually specifying $3$-$6$ keypoints per object on a reference image. During training or inference, we use visual foundation models~\cite{bommasani2021foundation-models} to extract the corresponding keypoints across variations in poses, objects, and scenes. These keypoints are then lifted to 3D and used as input for a diffusion policy~\cite{chi2025diffusion_policy}. To make informed design choices, we compare different extraction methods (\textit{image matching}, \textit{instance matching} and \textit{tracking}), foundation models and model architectures.

\begin{figure}[t] 
    \centering 
    \includegraphics[width=0.99\linewidth]{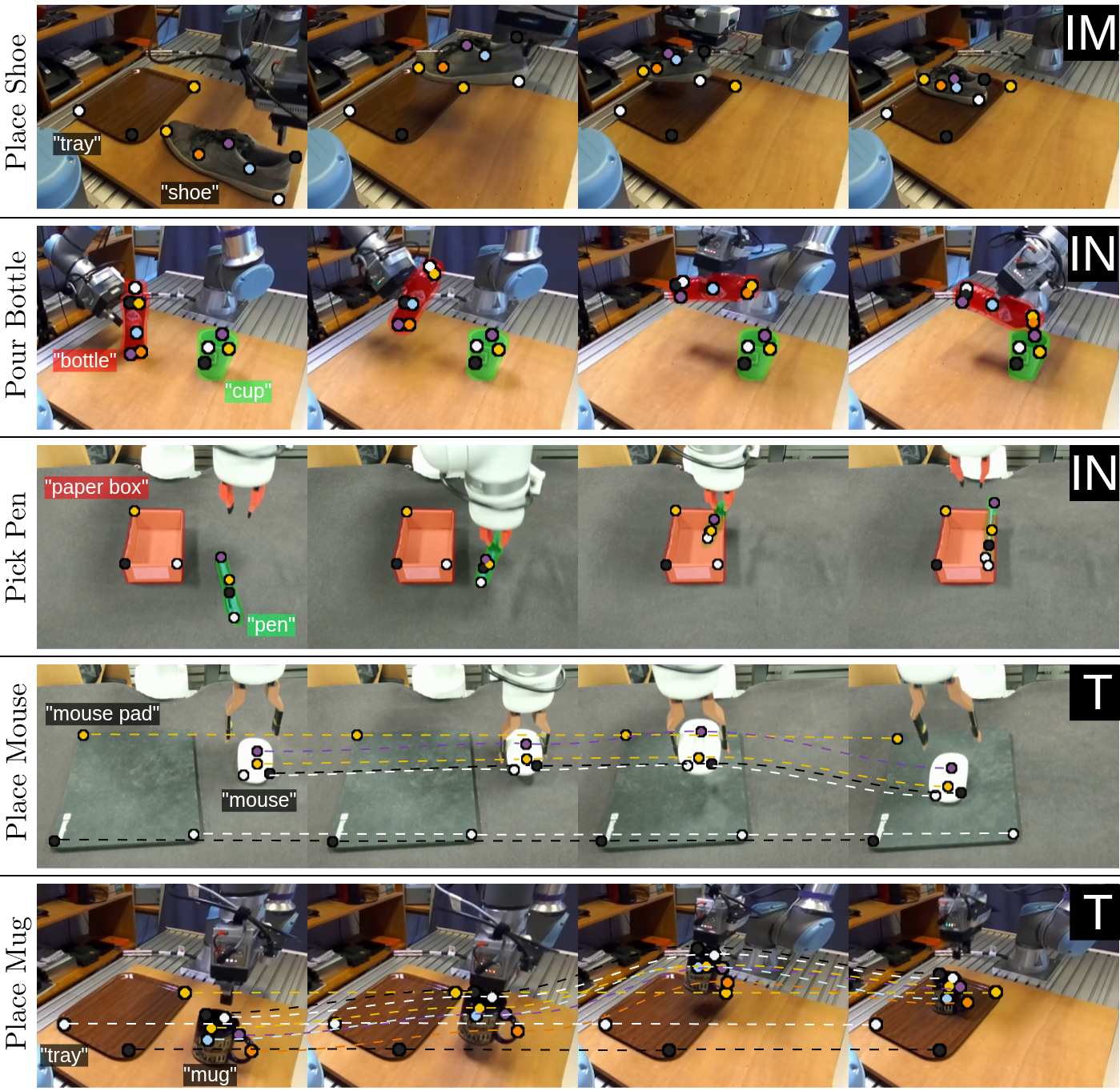} \caption{Successful real-world rollouts of our keypoint imitation learning policies. The top-right label indicates the keypoint extraction method used: IM=image matching, IN=instance matching, T=tracking. (cf. Section~\ref{sec:method})} 
    \label{fig:first_page} 
    \vspace{-0.0cm}
    \vspace{-\baselineskip}
\end{figure}

We perform over $2000$ real-world rollouts\footnote{\url{kil-manipulation.github.io}}, spanning two robot platforms and five different tasks, and evaluating both in-distribution performance, generalization to unseen objects, and scene variations. KIL achieves a $75\%$ success rate, compared to $47\%$ for an RGB diffusion 
baseline, a gap that widens dramatically under scene variations (KIL $70\%$ vs.\ RGB $10\%$). KIL performs on par with S$^2$-diffusion~\cite{yang2025S2} ($73\%$), an object-centric baseline that uses estimated depth maps and segmentation masks. Different keypoint extraction methods and foundation models perform similarly, with performance varying from $73$\% to $85$\%. We also extend KIL to multi-instance tasks and illustrate how current foundation models struggle with large variations in object orientations.

Our contributions are summarized as follows:
\begin{itemize}

    \item A performant pipeline for KIL, based on an investigation of several design choices, including keypoint extraction methods, foundation models and policy architectures.
    \item A thorough assessment of its generalization capabilities and limitations, using over $2000$ real-world evaluations on two robot platforms.
    \item The first demonstration of KIL applied to tasks with multiple instances of the same object category, extending its scope.

\end{itemize}

\begin{figure*}[h]
    \centering
    \includegraphics[width=0.95\linewidth]{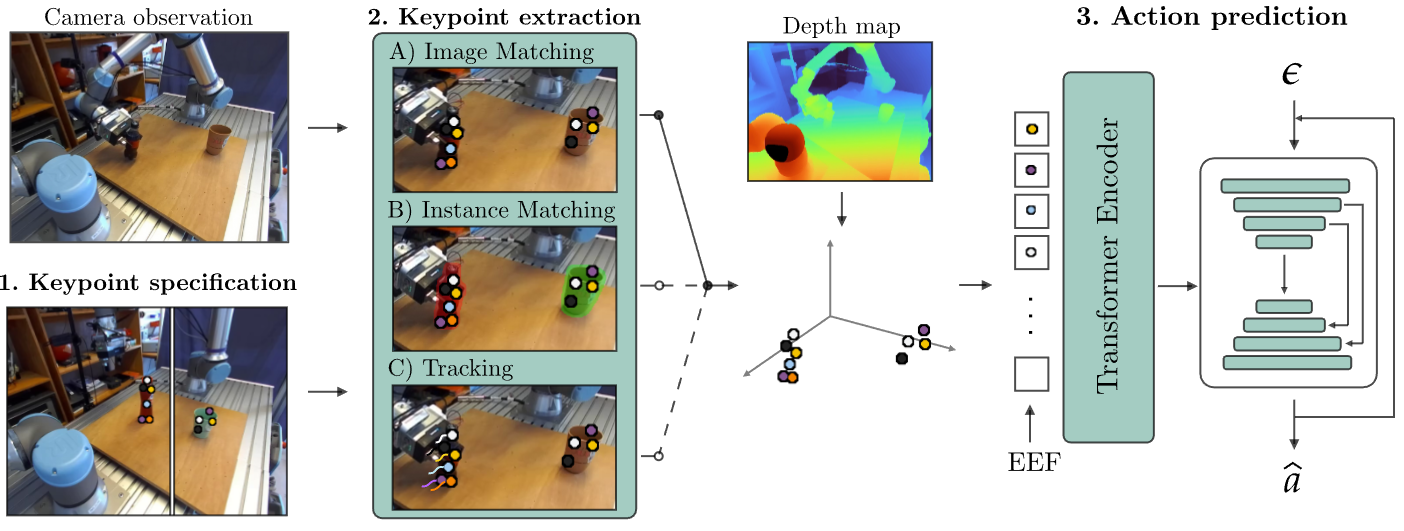}
    \caption{Overview of our three-stage keypoint imitation learning pipeline. 1) Keypoint references are manually annotated on a reference image for each object category. 2) Keypoints are extracted from each observation with visual foundation models and lifted to $3$D. 3) Keypoints are encoded by a transformer and used as input for a diffusion action head.}
    \label{fig:method-overview}
    \vspace{-0.0cm}
    \vspace{-\baselineskip}
\end{figure*}

\section{Related Work}
\label{sec:related-work}

\textbf{Keypoints for Robotic Manipulation} offer a compact, fine-grained and object-centric representation that has been used extensively~\cite{manuelli2019kpam,adriaens2025spill,vecerik2021s3k,chen2021unsupervised}.
Early works typically paired category-specific keypoint detectors with scripted policies~\cite{adriaens2025spill,manuelli2019kpam}. Others have focused on reducing labelling efforts~\cite{vecerik2021s3k,chen2021unsupervised}. More recently, visual foundation models~\cite{bommasani2021foundation-models} have enable 1-shot keypoint matching~\cite{amir2021ViTFeatures,tang2023dift,simeoni2025dinov3}, which can be used to build data-efficient robot manipulation systems~\cite{huang2025rekep,di2024KAT,wang2025skil}. 
In this work, we focus on using such keypoint extraction pipelines for imitation learning.



\textbf{Object-Centric Imitation Learning} aims to improve the efficiency and generalization of sensorimotor imitation learning, which has progressed rapidly~\cite{barreiros2025LBM-1,chi2025diffusion_policy}, by using alternative representations to RGB images, often using visual foundation models~\cite{bommasani2021foundation-models} to avoid task-specific labeling. 

S$^2$-diffusion~\cite{yang2025S2} combines depth maps and segmentation images to provide a geometric and object-centric representation. GROOT~\cite{zhu2023learning} and DP3~\cite{ze20243d} construct pointcloud representations while VIOLA~\cite{zhu2023viola} derives object-centric proposals from pretrained models.
Rana et al.~\cite{rana2024learning} use oriented affordance frames and 6D object poses to boost generalization.
In this work, we use both standard diffusion policy~\cite{chi2025diffusion_policy} with RGB images and S$^2$-diffusion~\cite{yang2025S2} as baselines for KIL.

\textbf{Visual Foundation Models for Keypoint Imitation Learning} have shown to improve in-distribution performance and generalization to unseen objects and scenes~\cite{Levy2025P3PO,di2024KAT,wang2025skil}. For keypoint specification, methods range from minimal human annotation on a single frame~\cite{haldar2025pointpolicy,Levy2025P3PO} to automatic distillation of task-driven points~\cite{vecerik2024robotap,zhang2025atk,wang2025skil,di2024KAT}. To extract keypoints from observations, these works either extract points from each frame via semantic descriptor matching~\cite{wang2025skil,di2024KAT} using various foundation models~\cite{tang2023dift,simeoni2025dinov3,heinrich2025radiov2.5}, or combine matching with explicit point-tracking~\cite{vecerik2024robotap,haldar2025pointpolicy,Levy2025P3PO,zhang2025atk}. 
Finally, the control policies vary widely, encompassing visual servoing primitives~\cite{vecerik2024robotap}, supervised imitation networks~\cite{zhang2025atk,wang2025skil,Levy2025P3PO,haldar2025pointpolicy} and training-free, LLM-based in-context learning, which outperforms learned policies in ultra-low data regimes (e.g., $10$ demos) but loses this advantage as the number of demonstrations increases~\cite{di2024KAT}. 

In this paper, we consolidate these approaches and conduct an in-depth study of keypoint imitation learning, comparing a.o. keypoint extraction strategies, foundation models, and policy designs.


\section{Method}
\label{sec:method}

Our method for keypoint imitation learning consists of three stages, illustrated in Figure~\ref{fig:method-overview}. First, a human operator specifies the semantic keypoints of interest by annotating a reference image for each object category (Section~\ref{sec:method-representation-specification}). Second, these keypoints are extracted from every observation, for both the demonstrations and during inference (Section~\ref{sec:method-representation-extraction}). Finally, the resulting sequence of $3$D keypoints is encoded by a transformer and used to condition a diffusion policy for action prediction (Section~\ref{sec:method-policy-learning}).



\subsection{Specifying the Keypoint Representation}
\label{sec:method-representation-specification}
As in~\cite{Levy2025P3PO,haldar2025pointpolicy}, we manually annotate three to six $2$D keypoints on a reference image selected from a single demonstration, for each object category the robot interacts with. We follow three heuristics when selecting keypoints: minimize occlusions and spread keypoints across the object; place a keypoint near any task-critical contact point (e.g.,\ the heel of a shoe); and maintain a small distance from object borders. 

\subsection{Representation Extraction}
\label{sec:method-representation-extraction}
Given the specification of the semantic keypoints, in the form of reference images with reference keypoints, we need to extract those keypoints across all initial states and for each timestep, both for training and inference.  We do this by first estimating the $2$D positions of the semantic keypoints and then lifting them to $3$D using the intrinsics and extrinsics of the calibrated RGB-D camera, as in~\cite{wang2025skil,Levy2025P3PO,di2024KAT}. 

There are different ways to extract the $2$D positions of the semantic keypoints in each frame. In this work, we consider the following three fundamental approaches:

\textbf{Image Matching (IM)}
Given the reference images and keypoints, we use a visual foundation model to compute a descriptor for each keypoint in the reference image. For each observation frame, we then compute these descriptors for every pixel in the image and use the cosine similarity with the reference descriptor of each keypoint to find the best match, as in~\cite{wang2025skil,di2024KAT}.

\textbf{Instance Matching (IN)}
This approach extends image matching by first running an open-vocabulary instance detector~\cite{carion2025sam3} (given a prompt for each category) to obtain $N$ instance masks per category, then restricting the descriptor-based cosine-similarity search to each object mask. 

This is a natural extension of the \textit{image matching} approach to handle multiple instances of the same category and potentially improve the matching accuracy. On the other hand, if the instance detector suffers from false positives or false negatives, this will impact the extracted keypoints significantly. An additional downside is the increased computational cost, as two foundation models must run for each observation. 

\textbf{Tracking (T)}
In the third and final approach, we use a dedicated tracking model to extract the keypoints, as in~\cite{Levy2025P3PO,haldar2025pointpolicy,vecerik2024robotap}.
To initialize the tracker, we use the \textit{instance matching} method to extract keypoints from the initial observation of a demonstration or rollout. Using separate models for tracking and initial matching can boost performance, especially with (short) object occlusions. 

These three different approaches for extracting keypoints each have their (dis)advantages. We compare them extensively in our experiments and refer to Section~\ref{sec:experiments-extraction-comparison} for the results. In Section~\ref{sec:experiments-model-comparison}, we also compare different foundation models for each extraction method.

\subsection{Policy Learning}
\label{sec:method-policy-learning}

Given the success of diffusion policies for imitation learning~\cite{chi2025diffusion_policy}, we use a diffusion policy as action head to predict robot actions, combined with a transformer encoder for the keypoints, as in~\cite{wang2025skil}.


The encoder closely resembles a standard transformer encoder~\cite{vaswani2017attention}, but we use pre-layer normalization, GeLU activations and no dropout.
The keypoints and the proprioceptive state are each processed as separate tokens. To tokenize them, we project them to the transformer dimension with a learned layer and add sinusoidal positional embeddings to encode their semantics. The output tokens are mean-pooled into a single vector, following~\cite{wang2025skil}. We use a history of observations as input for the policy, each processed independently in this way; the resulting observation vectors are concatenated and used to condition the diffusion action head, which predicts the next chunk of actions.

The actions are encoded as $10$-dimensional vectors, representing the absolute EEF position, absolute rotation, and relative gripper state. The rotation is encoded as a six-dimensional vector representing the first two columns of the rotation matrix to make the actions continuous~\cite{chi2025diffusion_policy}.
The proprioceptive state is encoded as a $7$-dimensional vector representing the absolute position, absolute orientation using Euler angles, and the relative gripper state.
All inputs and actions are normalized to (-1,1) using min-max scaling. Unlike the original diffusion policy, we do not use Exponential Moving Averaging (EMA) during training.
Finally, we augment our data by adding zero-mean Gaussian noise to the proprioceptive state and keypoints and applying a random translation and rotation around the $z$-axis to the inputs and actions, as in~\cite{chisari2024flow-matching-pc-IL}. \appendixnote{We report the hyperparameters for the augmentations in Appendix~\ref{sec:appendix-implementation-details-and-hyperparameters} and have evaluated the impact of the augmentations on KIL performance in Appendix~\ref{sec:appendix-additional-experiments-augmentations}.}

\section{Experiments}
\label{sec:experiments}

With our experiments, we aim to answer the following questions about KIL.

\noindent \textbf{RQ-1} Does KIL boost generalization to unseen scenes and objects, compared to alternative methods? (Section~\ref{sec:experiments-generalization})

\noindent \textbf{RQ-2} How do the three keypoint extraction methods described in Section~\ref{sec:method-representation-extraction} perform? (Section~\ref{sec:experiments-extraction-comparison})

\noindent \textbf{RQ-3} Which visual foundation models are best suited for extracting the keypoint representations? (Section~\ref{sec:experiments-model-comparison})

\noindent \textbf{RQ-4} How should we encode the keypoints in the policy architecture? (Section~\ref{sec:experiments-encoder-comparison})

\noindent \textbf{RQ-5} Can we extend the use of keypoint imitation to tasks with multiple instances of the same object category? (Section~\ref{sec:experiments-multi-objects})


Next to answering these questions, we illustrate how current foundation models limit the performance of KIL under large object orientation variations in Section~\ref{sec:experiments-orientation}.

\subsection{Hardware Setup}
\label{sec:experiments-hardware}
Both hardware setups used in this work are shown in Figure~\ref{fig:robot_setup}.
The first setup uses a UR5e Robot with a Schunk EGK-$40$ gripper and a Zed2i RGB-D camera. We use a custom tactile fingertip~\cite{liu2025magtouch}, to limit the grasping force of the gripper.
The second setup consists of a Ufactory lite6 with an integrated gripper and an Orbbec Femto Bolt RGB-D camera.
In both setups, the scene cameras are placed as close as possible while having a view of the whole workspace. The cameras are opposing the robot at an inclination of approximately $45$ degrees, to limit occlusions by the gripper and self-occlusion of each object in the scene. A Meta Quest $3$ headset running Quest2ROS~\cite{welle2024quest2ros} is used for demonstration collection. 

Policy and teleoperation actions are generated at $10$Hz and then interpolated to $50$Hz before sending them to the robot, to ensure smooth tracking. 

\subsection{Tasks and Data Collection}
\label{sec:experiments-tasks}
To test the capabilities and limitations of KIL, we designed five tasks that require the robot to manipulate different object categories. The choice for these objects is largely motivated by prior work~\cite{Levy2025P3PO,wang2025skil,di2024KAT,huang2025rekep}. 
We collect $50$ demonstrations per task, taking about $1$ hour for each task.  This is slightly more than previous works~\cite{Levy2025P3PO,di2024KAT,wang2025skil}, but provides a practical trade-off between performance and data collection effort. We vary the position of the objects significantly during data collection, but, similar to~\cite{wang2025skil,di2024KAT}, we limit the variation in orientations to \(\pm 30^\circ\) around a canonical orientation (cf. Section~\ref{sec:experiments-orientation}). We collect demonstrations using two different objects, but without scene variations, as in~\cite{Levy2025P3PO,wang2025skil,di2024KAT}.

The tasks and success criteria are described below. These criteria are combined with safety measures and a time limit (approx. twice the time it takes to teleoperate the task, excluding inference time) to determine if a rollout is successful.

\textbf{Place Shoe}: The goal is to pick a randomly placed shoe by the heel and put it on a tray that is placed at a fixed location. The task is successful if this happens without moving the tray significantly and with the shoe oriented within $90$ degrees of the y-axis of the tray.

\textbf{Place Mug} This task is similar to the previous one, but now requires picking a randomly placed, upright mug by the rim and placing it on a tray with a fixed position. 

\textbf{Pour Bottle}: The robot needs to pick a plastic 
bottle, move it over a cup, and tilt it as if to pour the contents into the cup.  The positions of the bottle and cup are randomized across the workspace. Since both objects are rotationally symmetric around the z-axis, the orientation does not matter for this task. The task is successful if the bottle is tilted beyond horizontal and is held directly above the cup. If the bottle is dropped, the task is terminated.

\textbf{Pick Pen}: The goal is to pick up a randomly placed pen and drop it in a cardboard box placed at a fixed location. The task is successful if the pen falls inside or remains on top of the box, without dropping the pen or moving the box significantly.

\textbf{Place Mouse}:  The goal is to grasp a randomly positioned mouse from the sides and place it on top of a mouse pad placed at a fixed location. The task is successful if the mouse is placed on the computer pad, facing the scene camera with the scroll wheel, without moving the pad significantly or dropping the mouse.

\begin{figure}[t] 
    \centering 
    \includegraphics[width=0.99\linewidth]{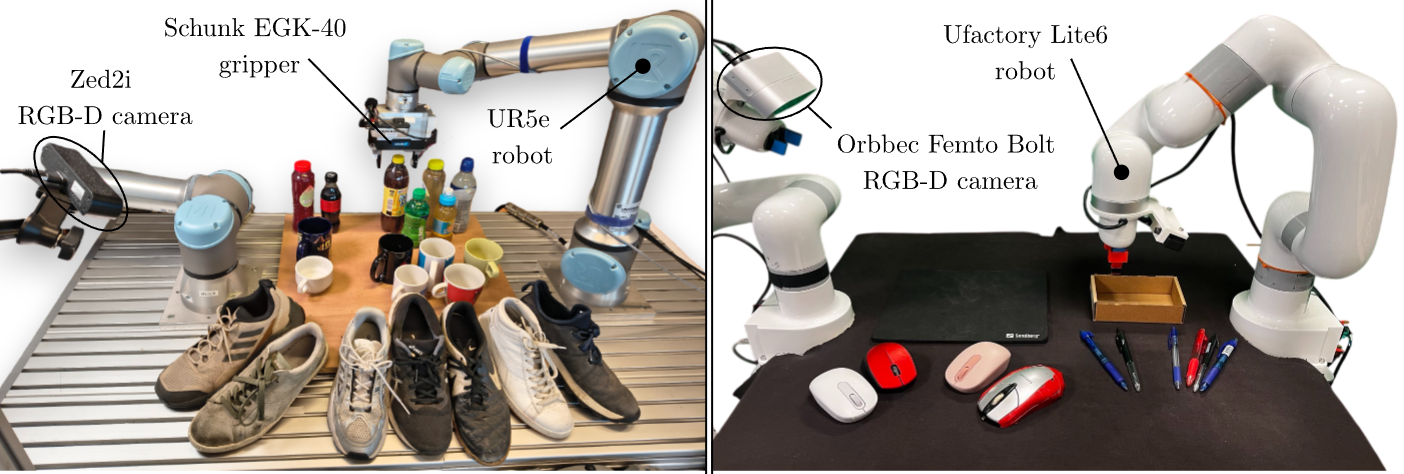} \caption{The $2$ robot setups and object sets used for each task (training objects grouped left of the evaluation objects for each task). } 
    \label{fig:robot_setup} 
    \vspace{-0.0cm}
    \vspace{-\baselineskip}
\end{figure}

An example of each task and its keypoint representation can be seen in Figure~\ref{fig:first_page}, while the objects used for data collection and evaluation are shown in Figure~\ref{fig:robot_setup}. \appendixnote{The keypoint annotations (cf. Section~\ref{sec:method-representation-specification}) for each task are shown in Appendix~\ref{sec:appendix-keypoint-references}, while an overlay of the initial frames of the demonstrations for each task are shown in Appendix~\ref{sec:appendix-demo-configurations}.}

\subsection{Baselines}
\label{sec:experiments-baselines}

We compare KIL against two baselines. The first is a standard RGB-based diffusion policy~\cite{chi2025diffusion_policy}. This uses the same diffusion action head as the keypoint policies, but now conditioned on the output of an image encoder that processes the RGB observation, resized to $320\times240$. We make one change to the encoder: we do not replace the global average pooling with a spatial softmax.
The second baseline is S$^2$-diffusion~\cite{yang2025S2}, which uses foundation models to generate a relative, normalized depth image and segmentation mask of the RGB image, using the policy architecture from~\cite{chi2025diffusion_policy}. 

\begin{table*}[h!] 
    \centering
    \renewcommand{\arraystretch}{1.2}
    \small 
    \setlength{\tabcolsep}{6pt} 
    \caption{Comparing keypoint extraction methods for KIL against baselines. IM=\textit{IMage matching}, IN=\textit{INstance matching},T=\textit{Tracking}. Each method is evaluated on $3$ sets of initial configurations, measuring in-distribution performance (IN), generalization to unseen objects (OBJ) and scene variations (S).}
    
    \begin{tabularx}{\textwidth}{l *{18}{Y} c c} 
        \toprule
        \multirow{2}{*}{\textbf{Method}} & 
        \multicolumn{3}{c}{\textbf{place shoe}} & 
        \multicolumn{3}{c}{\textbf{place mug}} & 
        \multicolumn{3}{c}{\textbf{pour cup}} & 
        \multicolumn{3}{c}{\textbf{pick pen}} & 
        \multicolumn{3}{c}{\textbf{place mouse}} & 
        \multicolumn{3}{c}{\textbf{ Task Avg. (\%)}} &
        \multirow{2}{*}{\textbf{Avg. (\%)}} &
        \multirow{2}{*}{\textbf{CI (\%)}} \\
        
        \cmidrule(lr){2-4} \cmidrule(lr){5-7} \cmidrule(lr){8-10} \cmidrule(lr){11-13} \cmidrule(lr){14-16} \cmidrule(lr){17-19}
        
        & \textbf{in} & \textbf{obj} & \textbf{s} & \textbf{in} & \textbf{obj} & \textbf{s} & \textbf{in} & \textbf{obj} & \textbf{s} & \textbf{in} & \textbf{obj} & \textbf{s} & \textbf{in} & \textbf{obj} & \textbf{s} & \textbf{in} & \textbf{obj} & \textbf{s} & & \\
        \midrule
        RGB~\cite{chi2025diffusion_policy}           & 8  & 6 & 3 & 10 & 7 & 0  & 4 & 0 & 1 & 7 & 7 & 0 & 8  & 8 & 1 & 74\% & 56\% & 10\% & 47\% & {\scriptsize [38, 55]} \\
        S$^2$~\cite{yang2025S2}            & 9  & 9 & 5 & 8  & 9 & 10 & 6 & 3 & 5 & 7 & 7 & 5 & 10 & 9 & 8 & 80\% & 74\% & 66\% & 73\% & {\scriptsize [66, 80]} \\
        KIL (IM)            & 10 & 6 & 9 & 10 & 8 & 8  & 7 & 4 & 5 & 7 & 8 & 4 & 10 & 8 & 9 & 88\% & 68\% & 70\% & 75\% & {\scriptsize [68, 82]} \\
        KIL (IN) & 10 & 6 & 5 & 10 & 7 & 10 & 1 & 4 & 2 & 8 & 9 & 6 & 9  & 9 & 9 & 76\% & 70\% & 64\% & 70\% & {\scriptsize [62, 77]} \\
        KIL (T) & 9  & 6 & 6 & 9  & 9 & 9  & 8 & 8 & 8 & 6 & 8 & 7 & 6  & 6 & 7 & 76\% & 74\% & 74\% & 75\% & {\scriptsize [67, 81]} \\
        \bottomrule
    \end{tabularx}
    \label{table:single-object-main}
    \vspace{-0.0cm}
    \vspace{-\baselineskip}
\end{table*}

\subsection{Experimental Setup}
\label{sec:experiments-setup}
\textbf{Implementation Details}
For \textit{image matching (IM)}, we use RADIOv2.5-B~\cite{heinrich2025radiov2.5} as feature extractor. For \textit{instance matching (IN)}, we use SAM3~\cite{carion2025sam3} to predict instance masks and DIFT~\cite{tang2023dift} with Stable Diffusion 2.1~\cite{rombach2022stable-diffusion} as feature extractor.
We use an ensemble size of $2$ for DIFT to balance inference cost and performance. For \textit{tracking (T)}, we use CoTracker3~\cite{karaev2025cotracker3} with the online modification from~\cite{Levy2025P3PO}; the initial frame is matched using SAM3 and DIFT with an ensemble size of $8$ to reduce variance due to the random noise initialization of DIFT. All images are resized to $640\times480$. We use bilinear interpolation to upscale the output of all feature extractors to the original size, as in~\cite{wang2025skil,huang2025rekep}.

Our keypoint transformer encoder consists of $4$ layers, using a token size of $128$ and an MLP size of $512$. The number of learnable parameters is approximately $1M$. As action head, we use the Unet from~\cite{chi2025diffusion_policy} and keep most hyperparameters at their default values. The observation history is set to $2$.
For all policies, we predict $16$ actions and execute the next $8$ actions before predicting the next action chunk.

For keypoint-based policies, we halve the number of channels in the Unet, since we did not observe improvements by using a larger action head, in line with~\cite{lin2025imitation_scaling_laws}. We also use a larger batch size of  $128$, compared to $32$ for the image-based baselines. All policies use a learning rate of $1e^{-4}$ with a cosine scheduler, as in~\cite{chi2025diffusion_policy}.
We train image-based baselines for $100k$ steps, and KIL for $250k$ steps. 
\appendixnote{Additional details on the implementation, hyperparameters and inference latencies are reported in Appendix~\ref{sec:appendix-implementation-details-and-hyperparameters}.}

\textbf{Evaluation protocol} 
We perform $30$ rollouts per task: $10$ with in-distribution initial configurations, $10$ with unseen objects (using $3-5$ objects per task, shown in Figure~\ref{fig:robot_setup}), and $10$ with scene variations, which include distractors and changing the background color of the surface. 
We use the same initial scene configurations \appendixnote{(cf. Appendix~\ref{sec:appendix-eval-configurations})} for the different policies to ensure fairness by overlaying reference images on the current scene and manually positioning all objects accordingly. Additionally, we control lighting conditions and background elements to limit distribution shifts between evaluations.

\textbf{Statistical Analysis} 
Next to providing point estimates of the aggregated success rate or task completion, we provide 95\% confidence intervals (CI) to quantify uncertainty and assess statistical significance. For binary success results, we assume a binomial distribution, as in ~\cite{barreiros2025LBM-1} 
and report Clopper-Pearson CIs for the average success rate. For task completion scores, we
rely on the central limit theorem ($N=60$) to assume a normal distribution of the average completion rate, as in~\cite{barreiros2025LBM-1}. We report the standard Wald CI $\hat{\mu} \pm 1.96\,\hat{\sigma}/\sqrt{N}$. \appendixnote{Finally, for tables that only include results aggregated over multiple tasks or evaluation scenarios, we report the full results in Appendix~\ref{sec:appendix-detailed-results}.}


\subsection{Comparing KIL Against Baselines}
\label{sec:experiments-generalization}
In this first experiment, we compare the performance of KIL against the two baselines, standard RGB-based diffusion and S$^2$ (\textbf{RQ-1}), across all five tasks described in Section~\ref{sec:experiments-tasks}.
The results are presented in Table~\ref{table:single-object-main}. KIL performs significantly better than the RGB-based diffusion policy overall, with the RGB policy achieving only $47\%$ success compared to $75\%$ for KIL with \textit{image matching}, the best-performing extraction method (cf.\ Section~\ref{sec:experiments-extraction-comparison}). While the RGB policy matches KIL on in-distribution evaluations and generalizes surprisingly well to unseen objects with varying appearance and geometry, it achieves only $10\%$ success on scene variations. S$^2$, by contrast, performs on par with KIL even under scene variations, which noticeably alter the depth maps relative to the demonstrations. Compared to prior work~\cite{di2024KAT,Levy2025P3PO,wang2025skil}, KIL achieves similar task performance, while our RGB policies perform slightly better, which we attribute to the larger number of training demonstrations.

Imprecise grasping was the most prominent failure mode across all policies: the robot nearly misses the object multiple times until a timeout occurs, or obtains a poor grasp that causes the rollout to fail later on, e.g.\ by dropping the object in a way that recovery is not possible. For KIL specifically, grasp failures occurred even when the extracted keypoints appeared accurate, suggesting insufficient policy generalization. Even in successful rollouts, KIL occasionally exhibited reduced grasping precision, which we attribute to variance and inaccuracies in the extracted keypoints.



\subsection{Comparing Keypoint Extraction Methods}
\label{sec:experiments-extraction-comparison}
We now compare the three methods for extracting keypoint representations discussed in Section~\ref{sec:method-representation-extraction}: \textit{image matching (IM)}, \textit{instance matching (IN)} and \textit{tracking (T)} (\textbf{RQ-2}).
In Table~\ref{table:single-object-main}, we find that all three methods perform similarly, achieving success rates of $75\%$, $70\%$ and $75\%$, averaged over the five tasks and different evaluation settings.

Empirically, \textit{matching}-based approaches for keypoint extraction suffer more from occlusions than the \textit{tracking} approach. At the same time, \textit{tracking} is more prone to drifting over time if objects are occluded for too long, or when the extracted keypoints of the first frame are inaccurate. 
We illustrate both keypoint failure modes in Figure~\ref{fig:failures}. 

Since we found no noticeable differences in performance between the extraction methods, we have used \textit{image matching}, which is simpler and computationally less expensive, for all experiments in this paper, unless stated otherwise.

\subsection{Comparing Feature Models for Keypoint Matching}
\label{sec:experiments-model-comparison}

To find the most suitable keypoint feature model for KIL (\textbf{RQ-3}), we compare three foundation models across different keypoint extraction methods on two tasks: \textit{Place Shoe} and \textit{Place Mug}. The three models are RADIOv2.5-B~\cite{heinrich2025radiov2.5}, DIFT~\cite{tang2023dift} and DINOv3-B~\cite{simeoni2025dinov3}.

Table~\ref{tab:model-comparison} shows that RADIOv2.5-B ($85\%$) performs better than DINOv3-B ($73\%$) and DIFT ($62\%$) for \textit{image matching}. We noticed that, although DIFT produces very accurate matches without occlusions, it is more prone to having incorrect matches, especially when the actual object is partially occluded and with scene variations. Using either \textit{instance matching} or \textit{tracking} reduces the impact of this issue, as can be seen in Table~\ref{tab:model-comparison}.
Finally, for \textit{instance matching}, and \textit{tracking}, we find no significant differences between RADIOv2.5-B and the DIFT models with ensemble sizes of $2$ and $8$. 

Based on these results, we have used RADIOv2.5-B for \textit{image matching}, DIFT$_2$ for \textit{instance matching} and DIFT$_8$ for \textit{tracking} in this paper, unless stated differently.




\begin{table}[h]
    \centering
    \renewcommand{\arraystretch}{1.2}
    \setlength{\tabcolsep}{5pt} 
    \caption{Comparison of keypoint feature models for the different keypoint extraction methods.}
    \begin{tabularx}{\linewidth}{l l Y Y c c}
        \toprule
        \makecell[l]{\textbf{Extraction} \\ \textbf{Method}} & \makecell[l]{\textbf{Feature} \\ \textbf{Model}} & \makecell[c]{\textbf{place} \\ \textbf{shoe} \\ \textbf{(/30)}} & \makecell[c]{\textbf{place} \\ \textbf{mug} \\ \textbf{(/30)}} & \makecell{\textbf{Avg.} \\ \textbf{(\%)}} & \makecell{\textbf{CI} \\ \textbf{(\%)}} \\
        \midrule
        \multirow{3}{*}{\textit{Image}} & \textbf{RADIOv2.5-B}  & 25 & 26 & 85\% & {\scriptsize [73, 93]} \\
        & DIFT$_2$  & 13 & 24 & 62\%  & {\scriptsize [48, 74]} \\
        & DINOv3-B & 20 & 24 & 73\%  & {\scriptsize [60, 84]} \\
        \midrule
        \multirow{2}{*}{\textit{Instance}} & RADIOv2.5-B & 19 & 27 & 77\% & {\scriptsize [64, 87]} \\
        & \textbf{DIFT$_2$} & 21 & 27 & 80\% & {\scriptsize [68, 89]} \\
        \midrule
        \multirow{3}{*}{\textit{Tracking}} &RADIOv2.5-B & 20 & 24 & 73\% & {\scriptsize [60, 84]} \\
        & DIFT$_2$ & 23 & 26 & 82\% & {\scriptsize [70, 90]} \\
        & \textbf{DIFT$_8$} & 21 & 27 & 80\% & {\scriptsize [68, 89]} \\
        \bottomrule
    \end{tabularx}
\label{tab:model-comparison}
\end{table}

\subsection{Comparing Keypoint Encoders}
\label{sec:experiments-encoder-comparison}

In this section, we aim to find the most suitable way to encode the keypoints before passing them to the diffusion action head (\textbf{RQ-4}). We compare our transformer encoder, described in Section~\ref{sec:method-representation-extraction}, against two alternatives. The first forgoes any encoder and simply concatenates the keypoints to the proprioceptive state of the diffusion policy, as in~\cite{di2024KAT}. The second uses the same transformer encoder but augments each keypoint's $3$D position with the cosine similarity against all keypoint feature descriptors, as in~\cite{wang2025skil}.

The results are shown in Table~\ref{tab:encoder-comparison}. We find that using no encoder performs on par with our transformer encoder, both achieving $85\%$ success rate. Appending cosine similarities to the keypoint $3$D positions, however, degrades performance dramatically to $7\%$. We hypothesize that, in combination with the geometric augmentations used in our pipeline, the network overfits on the similarity scores, which carry spurious information that is easier to memorize than the augmented $3$D positions. Removing augmentations substantially recovers performance for the similarity-based variant to $65\%$, but this is still below the performance we obtained without similarity scores. Beyond this negative interaction with augmentations, we also found the cosine similarity scores to be poorly calibrated across keypoints and scenes. Based on the results from this experiment, we have used the transformer encoder without similarities for all experiments in this paper, unless stated otherwise.
\begin{table}
    \centering
    \renewcommand{\arraystretch}{1.2}
    \setlength{\tabcolsep}{5pt}
    \caption{Encoder Comparison. Evaluated on the \textit{Place Shoe} and \textit{Place Mug} tasks.}
    \begin{tabularx}{0.6\linewidth}{l Y c}
        \toprule
        \textbf{Encoder} & \textbf{Avg.} & \textbf{CI (\%)} \\
        \midrule
        None              & 85\% & {\scriptsize [73, 93]} \\
        \textbf{ours}     & 85\% & {\scriptsize [73, 93]} \\
          \hspace{.3em} + sim & 7\% & {\scriptsize [2, 16]} \\
            \hspace{1.2em} - aug & 65\% & {\scriptsize [52, 77]} \\
        \bottomrule
    \end{tabularx}
    \label{tab:encoder-comparison}
    \vspace{-0.0cm}
    \vspace{-\baselineskip}
\end{table}





\subsection{Multi-Object Tasks}
\label{sec:experiments-multi-objects}
    
        
        
A limitation of using \textit{image matching} with multiple objects of the same category is that the keypoints might not be matched on each object, as multiple pixels with the highest similarity scores to a reference keypoint could be located on the same object, thus resulting in incomplete keypoint representations.
In this section, we extend KIL to such tasks (\textbf{RQ-5}) by extending the tasks from Section~\ref{sec:experiments-tasks} to make the robot handle two objects, instead of one.
During data collection, we have the robot handle the object closest to the camera first, to limit occlusions.

Model architecture, parameters, and keypoint references remain the same, with three exceptions:
(i) To account for the increased length of the tasks, we increase the number of training steps to $150K$ for all image-based baselines and $350K$ for KIL. 
(ii) Some changes are made to the extraction methods: For \textit{image matching}, we add a second reference image and keypoints to the configuration, allowing to extract keypoints for up to two objects. For \textit{instance matching}, we extract the two instances with the highest confidence and match keypoints on both masks. Should there be only one instance, we pad the list of keypoints with zeros, so that the total number of input tokens for the transformer encoder is consistent.
(iii) The token pooling strategy in the keypoint encoder is changed from mean-pooling to the output of a newly-added class token, concatenated with a maxpool over all other tokens, as in PointBERT~\cite{yu2022point-bert}. We made this change because the initial performance for the two-object tasks using mean-pooling was very poor, even though the keypoints were reasonably accurate. For the tasks from Section~\ref{sec:experiments-tasks}, changing the pooling strategy did not result in significant differences. \appendixnote{Our experiment with different pooling strategies is reported in Appendix~\ref{sec:appendix-additional-experiments-token-pooling}.}

\begin{table}
    \centering
    \renewcommand{\arraystretch}{1.2}
    \small 
    \setlength{\tabcolsep}{4pt}
    \caption{Task completion for tasks in which multiple instances of the same object category are handled. }
    
    \begin{tabularx}{0.95 \linewidth}{l Y Y c c}
        \toprule
        \textbf{Method} & \textbf{place 2 shoes (/30)} & \textbf{pick 2 pens (/30)} & \textbf{Avg. (\%)} & \textbf{CI (\%)} \\
        \midrule
        RGB~\cite{chi2025diffusion_policy}     & 13   & 15.5 & 48\% & {\scriptsize [37, 58]} \\
        S$^2$~\cite{yang2025S2}      & 14.5 & 20.5 & 58\% & {\scriptsize [47, 69]} \\
        KIL (IM) & 21.5 & 16.5 & 63\% & {\scriptsize [54, 73]} \\
        KIL (IN) & 21.5 & 15   & 61\% & {\scriptsize [50, 72]} \\
        KIL (T)  & 15.5 & 15.5 & 52\% & {\scriptsize [40, 63]} \\
        \bottomrule
    \end{tabularx}
    \label{table:2-object-experiments}
    \vspace{-0.0cm}
    \vspace{-\baselineskip}
\end{table}


The results of our experiments are shown in Table~\ref{table:2-object-experiments}. Note that instead of the final success rate, we now report task completion (defined as $0.5$ per correctly handled object) to better reflect model capabilities on these longer-horizon tasks~\cite{barreiros2025LBM-1}.  We find that KIL, irrespective of the extraction method, outperforms RGB ($48\%$) and performs similarly to S$^2$ ($58\%$). These results are consistent with the findings in Section~\ref{sec:experiments-generalization}, although the difference between RGB and KIL is smaller.


The strong performance of \textit{image matching (IM)} is surprising given the risk of incomplete representations. In practice, keypoints tended to match onto the most prominent object, which coincidentally aligned with the data collection strategy used in our experiments. We expect \textit{image matching} to break when the manipulated object is not the most visually prominent.

\begin{figure*}
    \centering
    \includegraphics[width=0.99\linewidth]{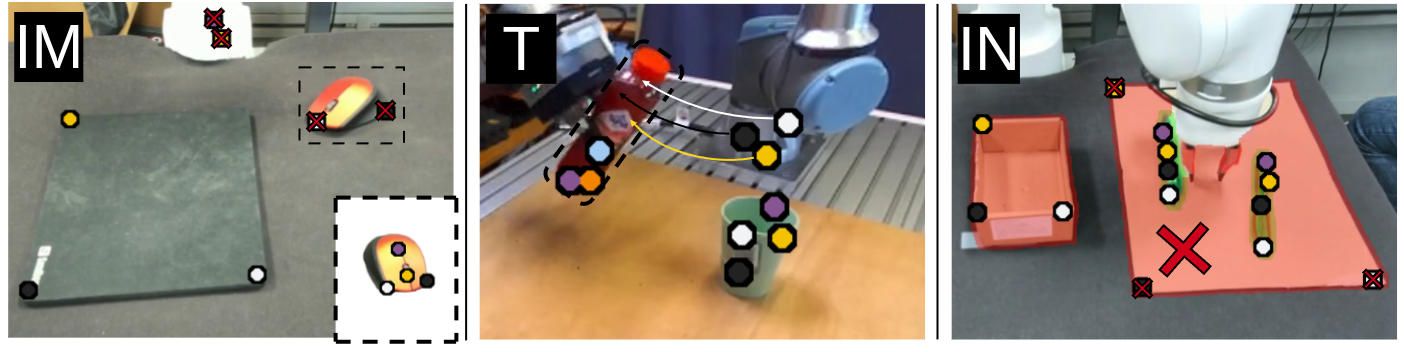}
    \caption{Failure examples for the keypoint extraction methods. Left (\textit{image matching}): matching assigns keypoints to incorrect parts of the scene rather than the intended object regions; the reference keypoints are shown in the bottom-right. Center (\textit{tracking}): drift causes some keypoints to snap onto a different object. Right (\textit{instance matching}): a false positive instance detection introduces additional keypoints. }
    \label{fig:failures}
    \vspace{-0.0cm}
    \vspace{-\baselineskip}
\end{figure*}

Next to the failure modes already described in Section~\ref{sec:experiments-extraction-comparison}, we noticed two additional failure modes in the multiple object experiments.
Firstly, for \textit{instance matching (IN)}, irrelevant scene elements can be falsely detected as a second object, causing failures if this did not happen during training. This is illustrated in Figure~\ref{fig:failures}.  Tuning the confidence threshold to avoid these false positives is non-trivial since a partially occluded true positive may score lower than a distractor.
Secondly, with \textit{tracking}, the tracker can snap onto another object of the same category when they move in front of each other, causing the policy to not see the second object. Per-frame matching approaches recover from this immediately once the second object is visible again.


\subsection{Robustness Against Rotations}
\label{sec:experiments-orientation}


A surprising limitation of current foundation models used for keypoint features is their poor robustness to large object orientation changes, together with their lack of calibration. Because of this, we restricted object orientations for all tasks in this paper, consistent with prior work~\cite{Levy2025P3PO,di2024KAT,wang2025skil}. To better illustrate this limitation, we now create task variants in which object poses span the full \(\pm 180^\circ\), for both the \textit{Place Mouse} and \textit{Pick Pen} tasks. We also impose an orientation constraint on the task success criteria: an execution is counted as successful only if the gripper grasps the object with an orientation that is within \(\pm 30^\circ\) of the orientation used consistently during teleoperation. 
\begin{table}
    \centering
    \renewcommand{\arraystretch}{1.2}
    \setlength{\tabcolsep}{5pt} 
    \caption{Illustration of performance drop when increasing the range of initial orientations for task objects. Evaluated on the \textit{Place Mouse} and \textit{Pick Pen} tasks.}
    \begin{tabularx}{0.95 \linewidth}{l Y Y c Y}
        \toprule
        & \multicolumn{2}{c}{\(\boldsymbol{\pm 180^\circ}\)} & \(\boldsymbol{\pm 30^\circ}\) & \(\boldsymbol{\Delta}\) \textbf{(30$\to$180)} \\
        \cmidrule(lr){2-3} \cmidrule(lr){4-4}
        \textbf{Method} & \textbf{Avg. (NS)} & \textbf{Avg.} & \textbf{Avg.} & \\
        \midrule
        RGB      & 70\% & 47\% & 52\% & $-10\%$ \\
        S$^2$    & 80\% & 70\% & 77\% & $-9\%$ \\
        KIL (IM) & 45\% & 35\% & 77\% & $-55\%$ \\
        KIL (T)  & 48\% & 48\% & 67\% & $-28\%$ \\
        \bottomrule
    \end{tabularx}
\label{tab:more-rotations-experiment}
\vspace{-0.0cm}
    \vspace{-\baselineskip}
\end{table}

Table~\ref{tab:more-rotations-experiment} reports the aggregated performance across all initial configurations for both tasks (Avg). To better expose the degradation of KIL under larger orientation variations, we also report an aggregate that excludes scene variation evaluations (\textit{Avg. (NS)} - no scene), since RGB performs poorly in this setting, which can partially mask the gap. We also report the performance on the corresponding \(\pm 30^\circ\) tasks, in addition to the relative difference between both ( $\Delta$ (30$\to$180)).
The results show a clear trend: keypoint-based methods degrade substantially when object orientation spans the full \(\pm 180^\circ\) range and correct gripper approach orientation is required, with KIL (IM) dropping by $55\%$ and KIL (T) by $28\%$. In contrast, RGB and S$^2$ are affected considerably less, dropping by only $10\%$ and $9\%$ respectively.

\section{Discussion and Limitations}

Similar to~\cite{di2024KAT,wang2025skil,Levy2025P3PO}, we found KIL to outperform RGB-based policies. However, as discussed in Section~\ref{sec:experiments-generalization}, we found the difference to be most significant under scene variations. 
Therefore, an interesting direction for future work is to compare KIL against methods that augment the demonstrations with image generation models~\cite{yu2023ROSIE} and test if the difference between KIL and RGB-based policies persists.

Unlike \cite{wang2025skil,Levy2025P3PO}, we found our object-centric baseline, S$^2$-diffusion, to perform on par with KIL. This raises the question of why the semantic granularity and compactness of keypoints does not translate into superior performance over representations such as masks or depth maps. A possible explanation is that the extraction pipelines (i.e. the foundation models) are more performant for those other representations.

We observed KIL to be less precise and degrade faster under occlusions and variations in object orientation than RGB-based policies. These are limitations of the current foundation models used to extract keypoints, rather than the concept of using keypoints for imitation learning. Potential directions to overcome them include end-to-end finetuning of the extraction pipeline or combining keypoints with other representations~\cite{chen2025history-keypoint-IL,cao2025COIL}, to combine their strengths.




We also identify several limitations of our work.
We did not consider the use of wrist cameras, which help to reduce occlusion and can boost generalization for RGB-based policies~\cite{hsu2022wrist-cam-policy-benefits}. We also used depth sensors in this work. Although we found their precision to be sufficient, the use of depth sensors limits the scope of tasks to a.o. objects that are not reflective or transparent. 
The performance of SAM3~\cite{carion2025sam3} for \textit{instance matching} depends a.o. on how discriminative the prompt is. For example, in the \textit{Pick Pen} task, we used "paper box" instead of "box" as prompt to improve performance. Such prompt engineering increases the time required for representation specification.
Finally, our investigation of KIL's generalization capabilities is limited to $2$-shot object generalization and $1$-shot scene generalization. A more extensive evaluation of how generalization scales with the number of demonstrated variations, as in~\cite{lin2025imitation_scaling_laws}, would provide a more complete picture. 

\section{Conclusion}

We present an in-depth evaluation of keypoint imitation learning (KIL). KIL outperforms RGB policies under scene variations but performs similarly within distribution and for unseen objects. Additionally, KIL performs on par with S$^2$-Diffusion, raising questions about the added value of keypoints compared to other representations. For keypoint extraction, our results show that \textit{image matching }performs similarly to the more complex  \textit{instance matching} and \textit{tracking} methods. We further showed that a dedicated transformer encoder improves performance, and also extended KIL to handle tasks with multiple instances of the same object category. Finally, we illustrated how KIL inherits poor robustness to large orientation changes as a limitation from the foundation models used for keypoint matching, in addition to their limited precision and ability to handle occlusions. 
Our findings clarify several design choices for KIL and confirm its potential for improved generalization. More research is needed to fully understand its characteristics and limitations, but this work provides a solid foundation for further exploration of keypoint representations for imitation learning.

\section*{ACKNOWLEDGMENTS}
\ifanonymous
Funding anonymized for double-blind review.
\else
T.L. and M.M. were funded by the European project euROBIN (grant No. 101070596); T.L. was also funded by FWO (grant No. 1S56024N).
\fi

We used LLMs (GPT5, Claude Sonnet/Opus 4.5) to assist with coding, paper formatting and proofreading.



\bibliographystyle{IEEEtran}
\bibliography{references}

\ifappendixincluded

\clearpage
\onecolumn 

\section*{APPENDIX}
\FloatBarrier 

\setcounter{subsection}{0}
\renewcommand{\thesubsection}{\Alph{subsection}}
\renewcommand{\thesubsubsection}{\Alph{subsection}.\arabic{subsubsection}}


\subsection{Additional Experiments}
\label{sec:appendix-additional-experiments}
In this appendix, we provide two additional experiments. In Appendix~\ref{sec:appendix-additional-experiments-augmentations}, we investigate the impact of the data augmentations described in Section~\ref{sec:method-policy-learning} on KIL performance. In Appendix~\ref{sec:appendix-additional-experiments-token-pooling}, we compare the performance of different token pooling strategies for KIL, for tasks with multiple object instances, as discussed in Section~\ref{sec:experiments-multi-objects}.
\subsubsection{Impact of Augmentations}
\label{sec:appendix-additional-experiments-augmentations}
We study the impact of data augmentations for KIL by selectively enabling the two main augmentations described in Section~\ref{sec:method-policy-learning}: adding \textit{Keypoint Noise} and applying a random \textit{Spatial Transform (ST)}. We measure performance on \textit{Pick Pen} and \textit{Place Mouse} using \textit{image matching} as keypoint extraction methods.

Table~\ref{tab:augmentation-comparison} reports results. We find that not using any augmentation ($42\%$) hurts performance, but do not observe meaningful differences between using \textit{KP Noise} ($80\%$), \textit{ST} ($70\%$) or both ($77\%$). For all experiments in this paper, we have used both augmentations. 

At some point, we also tested keypoint dropout augmentation, as in \cite{wang2025skil}, but found it to degrade performance significantly on some tasks, hence we did not use it in the experiments presented in this paper.

\begin{table*}
    \renewcommand{\arraystretch}{1.2}
    \begin{minipage}[t]{0.38\linewidth}
        \centering
        \setlength{\tabcolsep}{2pt}
        \caption{Impact of Augmentations on KIL. Evaluated on the \textit{Pick Pen} and \textit{Place Mouse} tasks.}
        \label{tab:augmentation-comparison}
        \begin{tabularx}{\linewidth}{l Y c}
            \toprule
            \textbf{Augmentations} & \textbf{Avg.} & \textbf{CI (\%)} \\
            \midrule
            None  & 42\% & {\scriptsize [29, 55]} \\
            Noise & 80\% & {\scriptsize [68, 89]} \\
            ST    & 70\% & {\scriptsize [57, 81]} \\
            \textbf{Noise + ST}   & 77\% & {\scriptsize [64, 87]} \\
            \bottomrule
        \end{tabularx}
    \end{minipage}%
    \hfill
    \begin{minipage}[t]{0.55\linewidth}
        \centering
        \small
        \setlength{\tabcolsep}{8pt}
        \caption{Comparison of different token aggregation strategies for the keypoint encoder. Evaluated on the 10 in-distribution configurations for each task.}
        \label{table:token-pooling-comparison}
        \begin{tabularx}{\linewidth}{l Y Y Y}
            \toprule
            \textbf{Method} & \textbf{2 shoes} & \textbf{2 pens} & \textbf{pour cup} \\
            \midrule
            ours & 2 & 5.5 & 7 \\
            ours (d=256) & 2.5 & 3.5 & 8 \\
            \textbf{ours (pointBERT)} & 8 & 7.5 & 9 \\
            none & 1.5 & 3 & 6 \\
            \bottomrule
        \end{tabularx}
    \end{minipage}
\end{table*}
\subsubsection{Token Pooling Strategy Comparison}
\label{sec:appendix-additional-experiments-token-pooling}
In this experiment we elaborate on the use of a different token aggregation strategy in Section~\ref{sec:experiments-multi-objects}. As stated before, we found that using pointBERT~\cite{yu2022point-bert} token aggregation (CLS token + maxpool) worked better than using mean-pooling, which we adopted from~\cite{wang2025skil} for these tasks with multiple object instances.

We hypothesize that as the number of (independent) keypoints increases, mean-pooling dilutes spatially relevant features by a factor of 
\(\frac{1}{N}\), reducing the accuracy of the representation under mean-pooling or forcing the network to learn a more complex mapping to compensate for the dilution.

In Table~\ref{table:token-pooling-comparison}, we compare mean pooling against pointBERT pooling. We also compare against widening the transformer encoder, thus providing more capacity to the tokens, or not using an encoder at all. We find that pointBERT  outperforms all other options for the two-object tasks, confirming the importance of using an appropriate token aggregation strategy.

We also applied pointBERT pooling to the single-object tasks, as illustrated by the comparison on the \textit{pour} task in Table~\ref{table:token-pooling-comparison}. We find that it performs similarly to mean-pooling.
We expect that the performance for the single-object tasks would in general improve slightly with pointBERT pooling. However, we expect it to boost performance for all keypoint-based methods equally, hence not affecting the conclusions of the experiments in this work. We recommend future works to avoid mean-pooling or to at least compare different aggregation strategies.

\subsection{Implementation Details and Hyperparameters}
\label{sec:appendix-implementation-details-and-hyperparameters}
In this section, we provide additional details on the implementation and hyperparameters used for the policies.

Table~\ref{tab:hparams} summarises the hyperparameters used for training, except for the augmentation hyperparameters, which are reported in Table~\ref{tab:augmentations}. 

Table~\ref{tab:inference-latency} reports the approximate inference latency for each method on an RTX 4090 with torch v2.9.1. We use bf16 precision for foundation model inference and fp32 for policy inference. We do not use torch.compile. The table reports both per-observation latency (preprocessing of each observation at 10Hz, if needed) and the per-inference-step latency (the full cost paid each time the policy is queried for a chunk of actions).

Finally, in Listing~\ref{lst:encoder}, we provide a minimal implementation of the encoder architecture used for the keypoint-based policies, as described in Section~\ref{sec:method-policy-learning}.

\begin{table}[h]
    \centering
    \renewcommand{\arraystretch}{1.3}
    \setlength{\tabcolsep}{8pt}
    \caption{Hyperparameters for KIL and baseline policies. All methods use a Diffusion Policy~\cite{chi2025diffusion_policy} action head.}
    \label{tab:hparams}
    \begin{tabularx}{\linewidth}{l Y Y}
        \toprule

        \textbf{Hyperparameter} & \textbf{RGB / S$^2$} & \textbf{KIL} \\
        \midrule
        Observation encoder & ResNet18 & Transformer \\
        Input resolution    & $320\times240$ & $640\times480$ \\
        Action Head Dims                          & [256,512,1024] & [128,256,512] \\
        Action Head kernel size   & \multicolumn{2}{c}{5} \\
        Noise scheduler   & \multicolumn{2}{c}{DDIM} \\
        Denoising steps   & \multicolumn{2}{c}{100 train /14 inference} \\
        Action prediction horizon ($T_p$)   & \multicolumn{2}{c}{16} \\
        Action horizon ($T_a$)                       & \multicolumn{2}{c}{8} \\
        Observation horizon ($T_o$) & \multicolumn{2}{c}{2} \\
        Peak LR       & \multicolumn{2}{c}{$10^{-4}$} \\
        LR schedule                            & \multicolumn{2}{c}{cosine} \\
        Optimizer & Adam & AdamW \\
        Weight decay & $10^{-6}$ & $10^{-5}$ \\
        Batch size          & 64 & 128 \\
        Training steps      & 100K & 250K \\
        Augmentations (Table~\ref{tab:augmentations})       & RandomCrop, Gaussian Noise, ColorJitter & KP Noise, ST, EEF Noise \\
        \bottomrule
    \end{tabularx}
\end{table}

\begin{table*}[h]
    \renewcommand{\arraystretch}{1.3}
    \begin{minipage}[t]{0.48\linewidth}
        \centering
        \setlength{\tabcolsep}{8pt}
        \caption{Augmentation hyperparameters. ST = Spatial Transform; KP = Keypoint; EEF = End-Effector.}
        \label{tab:augmentations}
        \begin{tabularx}{\linewidth}{l l X}
            \toprule
            \textbf{Method} & \textbf{Augmentation} & \textbf{Parameters} \\
            \midrule
            \multirow{3}{*}{RGB}
                & RandomCrop & $288\times216$ \\
                & Gaussian Noise & $\sigma = 12.5$ mm \\
                & ColorJitter & range$=(0.4, 0.4, 0.4, 0.1)$ \\
            \midrule
            \multirow{2}{*}{S$^2$}
                & RandomCrop & $288\times216$ \\
                & Gaussian Noise & $\sigma = 12.5$ px \\
            \midrule
            \multirow{3}{*}{KIL}
                & KP Noise & $\sigma = 1$ mm \\
                & ST & $\Delta t = 0.3$ m, $\Delta\theta_z = 0.6$ rad \\
                & EEF Noise & $\sigma_{\text{pos}} = 2$ mm, $\sigma_{\text{ori}} = 1\deg $, $\sigma_{\text{grip}} = 0.02$ \\
            \bottomrule
        \end{tabularx}
    \end{minipage}%
    \hfill
    \begin{minipage}[t]{0.49\linewidth}
        \centering
        \renewcommand{\arraystretch}{1.2}
        \setlength{\tabcolsep}{3pt}
        \caption{Inference latency (ms) of the different policies used in this work, measured on an NVIDIA RTX 4090 using bf16 precision for foundation model inference. We do not use torch.compile.}
        \label{tab:inference-latency}
        \begin{tabularx}{\linewidth}{l c Y}
            \toprule
            \textbf{Method} & \textbf{Per observation} & \textbf{Per inference step} \\
            \midrule
            RGB              & /  & $N_o \times$ ResNet18 + AH : 35 ms \\
            S2               &/ & $N_o \times$ (DAv2 + gSAM) + AH: 540 ms \\
            KIL (IM) & /    & $N_o \times$ RADIOv2.5-B + AH: 80 ms \\
            KIL (IN) & /  & $N_o \times$ (DIFT\textsubscript{2} + SAM3) + AH: 570 ms\\
            KIL (T)    & CoTracker3 : 59 ms           & AH : 29ms \\
            \bottomrule
        \end{tabularx}
    \end{minipage}
\end{table*}

\begin{figure}[h]
\noindent\begin{minipage}{\linewidth}
\begin{lstlisting}[language=Python, caption=Minimal implementation of the keypoint encoder, label=lst:encoder,
  basicstyle=\ttfamily\scriptsize, breaklines=false, columns=fullflexible]
  import torch, torch.nn as nn, numpy as np

  def sincos_embed(d, n):
      i = np.arange(d // 2) / (d / 2.0)
      e = np.outer(np.arange(n), 1.0 / 10000**i)
      return torch.from_numpy(np.concatenate([np.sin(e), np.cos(e)], axis=1)).float()
  
  class KeypointEncoder(nn.Module):
      def __init__(self, seq_len=9, kp_dim=3, ee_dim=7,d=128, nhead=8, num_layers=4, ff=512):
          super().__init__()
          self.seq_len = seq_len
          self.kp_dim  = kp_dim
          self.ee_dim  = ee_dim
          self.d       = d
  
          self.kp_emb = nn.Linear(kp_dim, d)
          self.ee_mlp = nn.Sequential(
              nn.Linear(ee_dim, d*2), nn.ReLU(),
              nn.Linear(d*2,   d*2), nn.ReLU(),
              nn.Linear(d*2,   d),   nn.Tanh(),
          )
          layer = nn.TransformerEncoderLayer(d, nhead, ff, dropout=0.0,activation='gelu', batch_first=True,norm_first=True)
          self.encoder = nn.TransformerEncoder(layer, num_layers, norm=None)
          self.register_buffer('pos', sincos_embed(d, seq_len + 1).unsqueeze(0))
  
      def forward(self, kp, ee):
          B, T = kp.shape[:2]
          kp_tokens = self.kp_emb(kp.reshape(B*T, self.seq_len, self.kp_dim)) + self.pos[:, :self.seq_len]
          ee_token  = self.ee_mlp(ee.reshape(B*T, self.ee_dim)).unsqueeze(1) + self.pos[:, self.seq_len:]
          tokens = torch.cat([kp_tokens, ee_token], dim=1)
          tokens = self.encoder(tokens)
          return tokens.mean(dim=1).reshape(B, T, -1)
\end{lstlisting}
\end{minipage}
\end{figure}






\subsection{Initial Scene Configurations of Demonstrations}
\label{sec:appendix-demo-configurations}
Figure~\ref{fig:demo-distributions} shows an overlay of the initial frames of the demonstrations for each task, illustrating the distribution of initial scene configurations used during data collection. As described in Section~\ref{sec:experiments-tasks}, we vary the position of objects significantly across the workspace, but limit orientation variation to $\pm 30^\circ$ around a canonical orientation. Demonstrations are collected using two different objects per task, without scene variations.

\begin{figure}[H]
    \centering
    \subfloat[Place Shoe]{\includegraphics[width=0.32\linewidth]{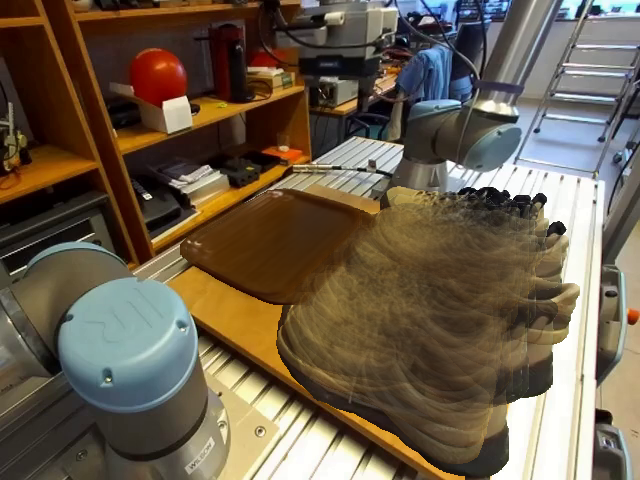}}%
    \hfill
    \subfloat[Place Mug]{\includegraphics[width=0.32\linewidth]{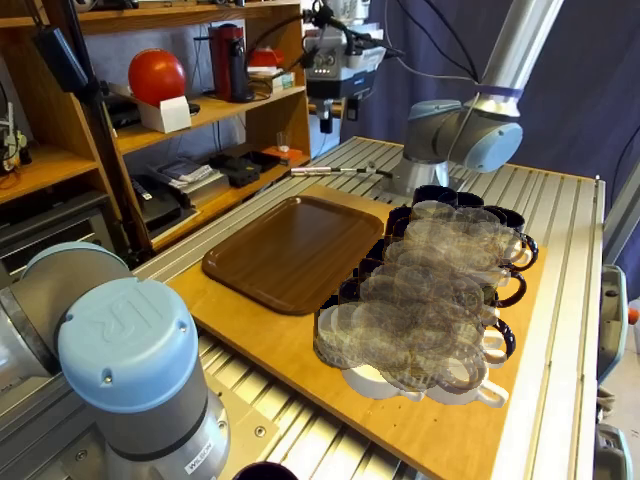}}%
    \hfill
    \subfloat[Pour Bottle]{\includegraphics[width=0.32\linewidth]{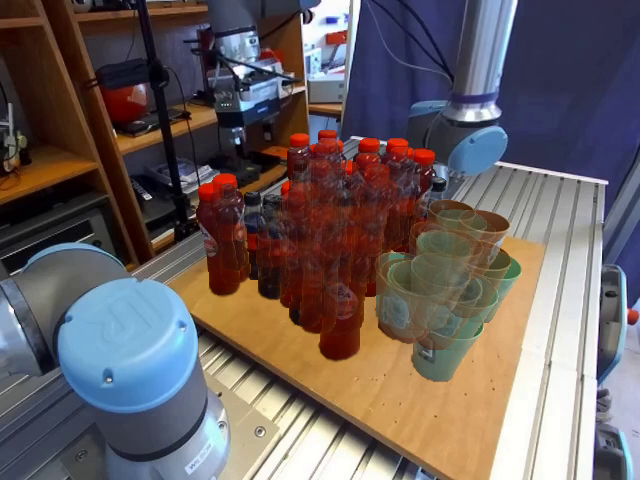}}\\
    \subfloat[Pick Pen]{\includegraphics[width=0.32\linewidth]{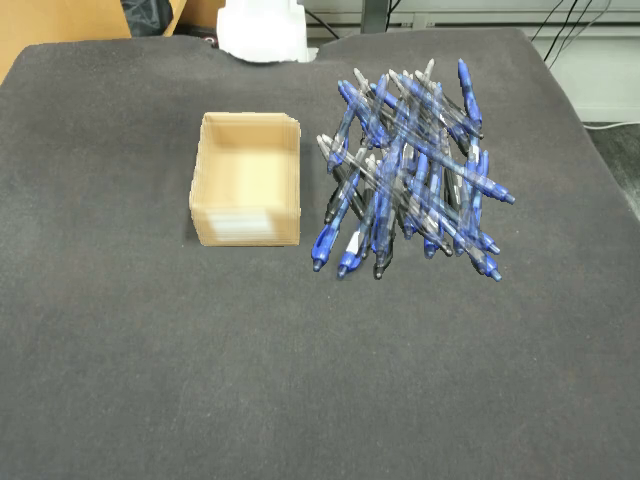}}%
    \hfill
    \subfloat[Place Mouse]{\includegraphics[width=0.32\linewidth]{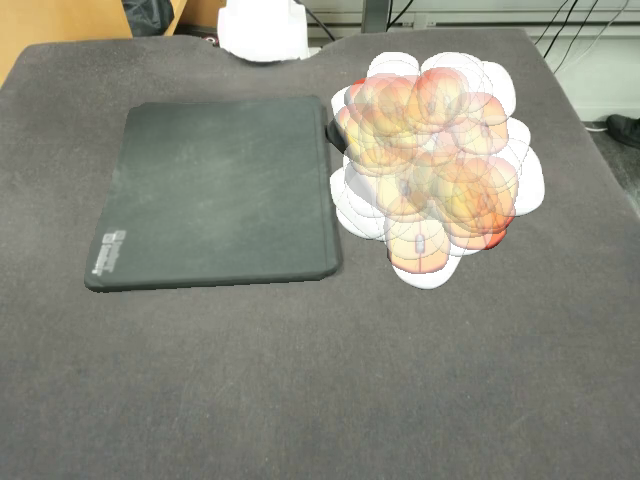}}%
    \hfill
    \subfloat{\phantom{\includegraphics[width=0.32\linewidth]{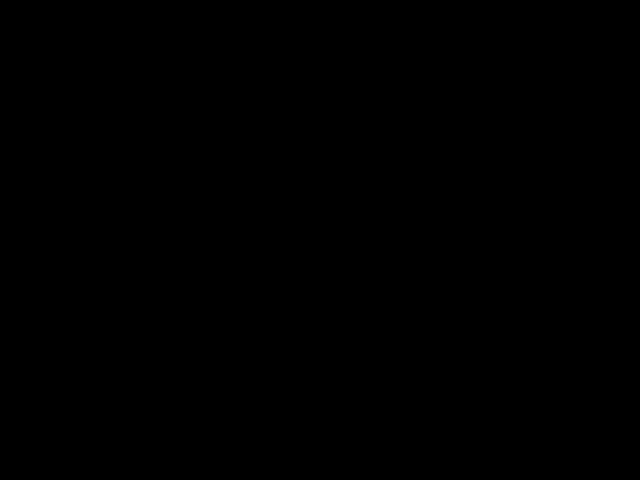}}}%
    \vspace{4pt}
    \caption{Scene configuration distribution of the demonstrations for each task.}
    \label{fig:demo-distributions}
\end{figure}

\subsection{Keypoint Representation Reference images}
\label{sec:appendix-keypoint-references}
Figure~\ref{fig:kp-configs} shows the keypoint reference images used for each task, as described in Section~\ref{sec:method-representation-specification}. Each image shows the reference frame with the annotated keypoints and text prompt used for instance detection (cf. Section~\ref{sec:method-representation-extraction}).

\begin{figure}[H]
    \centering
    \includegraphics[width=0.49\linewidth]{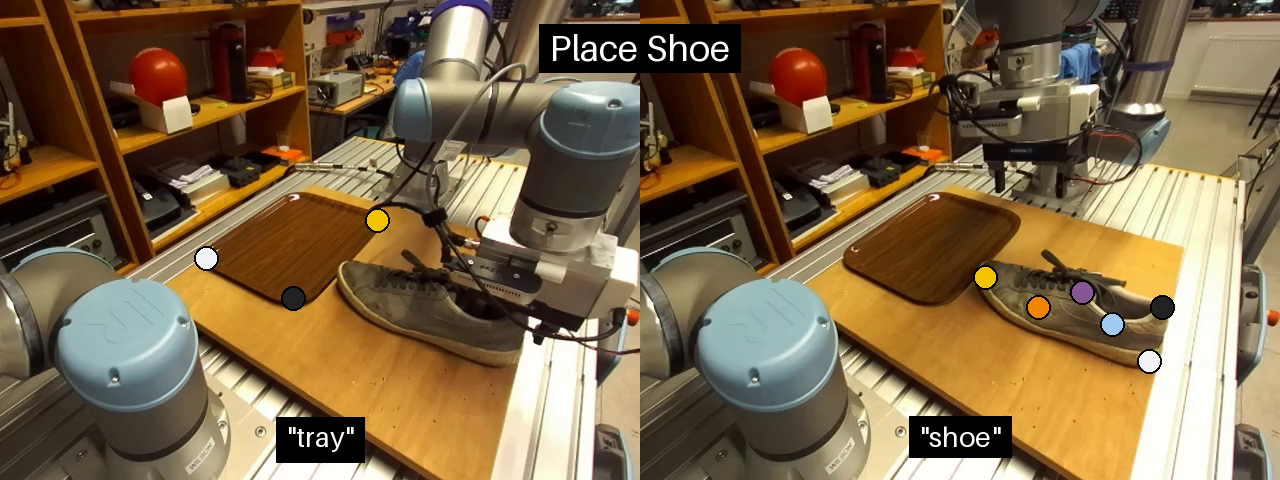}\hfill
    \includegraphics[width=0.49\linewidth]{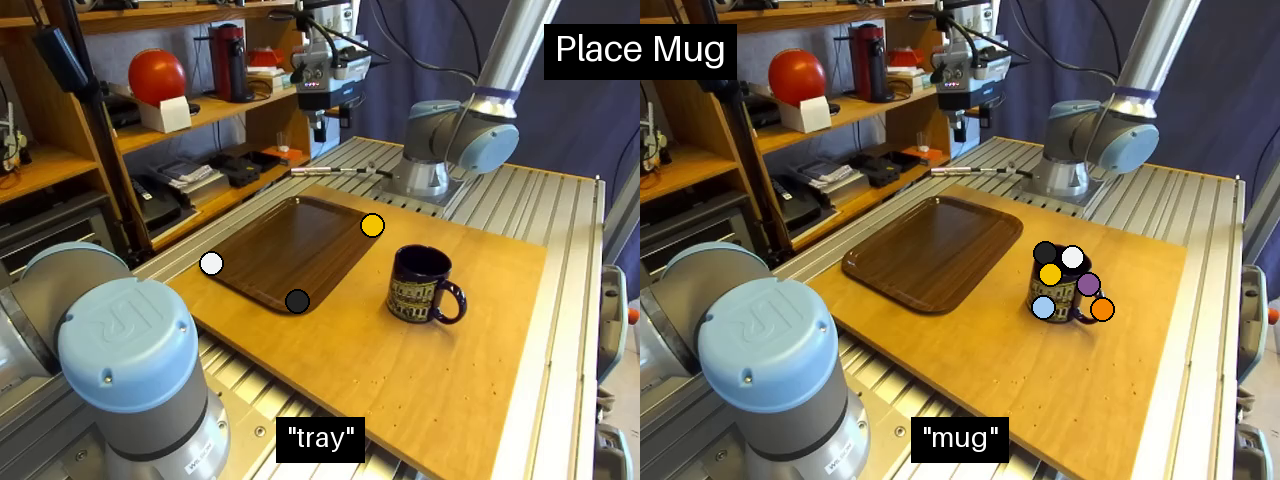}\\[8pt]
    \includegraphics[width=0.49\linewidth]{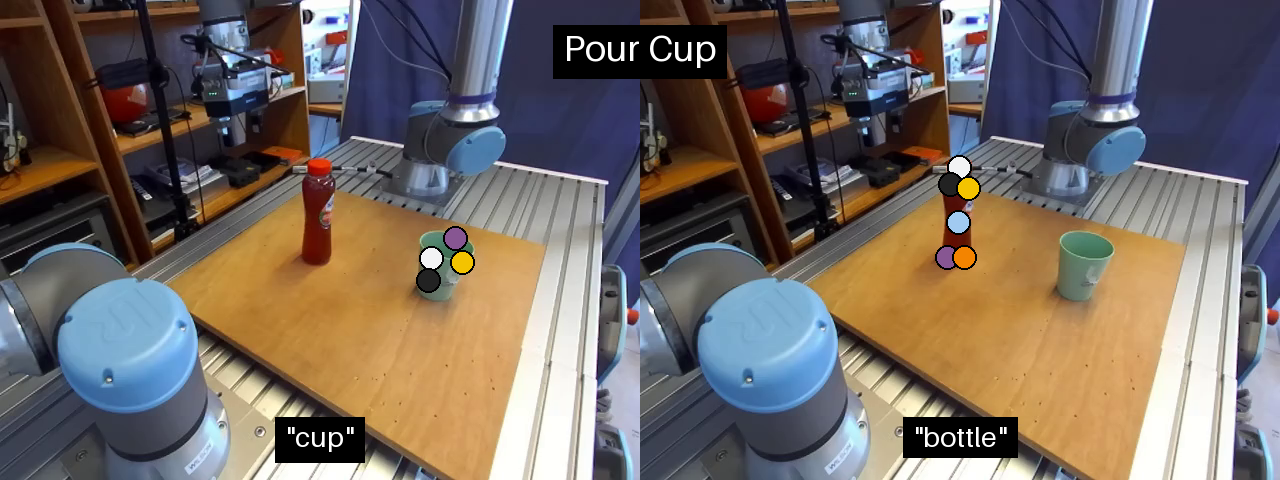}\hfill
    \includegraphics[width=0.49\linewidth]{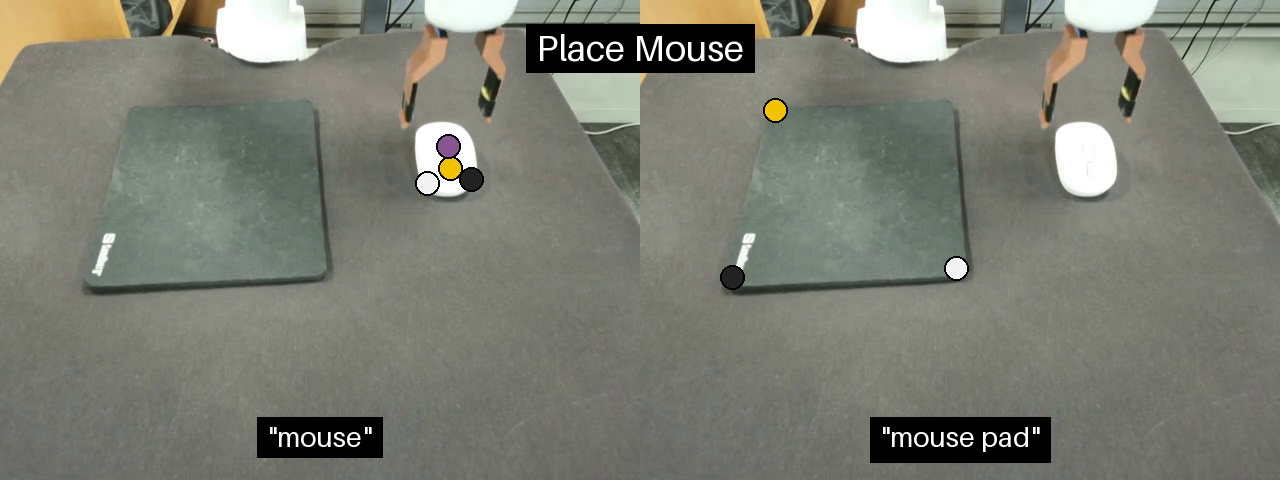}\\[8pt]
    \includegraphics[width=0.49\linewidth]{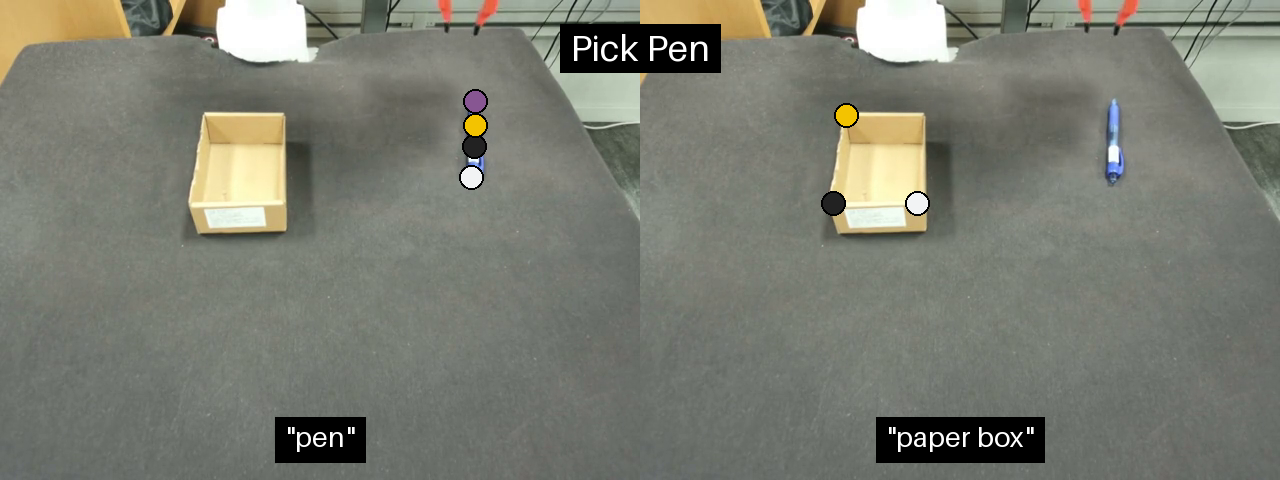}
    \caption{Keypoint reference images for each task, inluding text prompts for each object.}
    \label{fig:kp-configs}
\end{figure}

\subsection{Initial Scene Configurations for Evaluations}
\label{sec:appendix-eval-configurations}

Figure~\ref{fig:eval-configurations} shows the initial scene configurations used for evaluation. For each task, we measure in-distrubtion performance, generalization to unseen objects and generalization to scene variations. We use the same 10 initial scene configurations for all policies to ensure fairness.

\begin{figure*}[h]
    \centering
    \begin{minipage}[c]{0.03\linewidth}\centering\rotatebox{90}{Place Shoe}\end{minipage}%
    \begin{minipage}[c]{0.97\linewidth}\includegraphics[width=\linewidth]{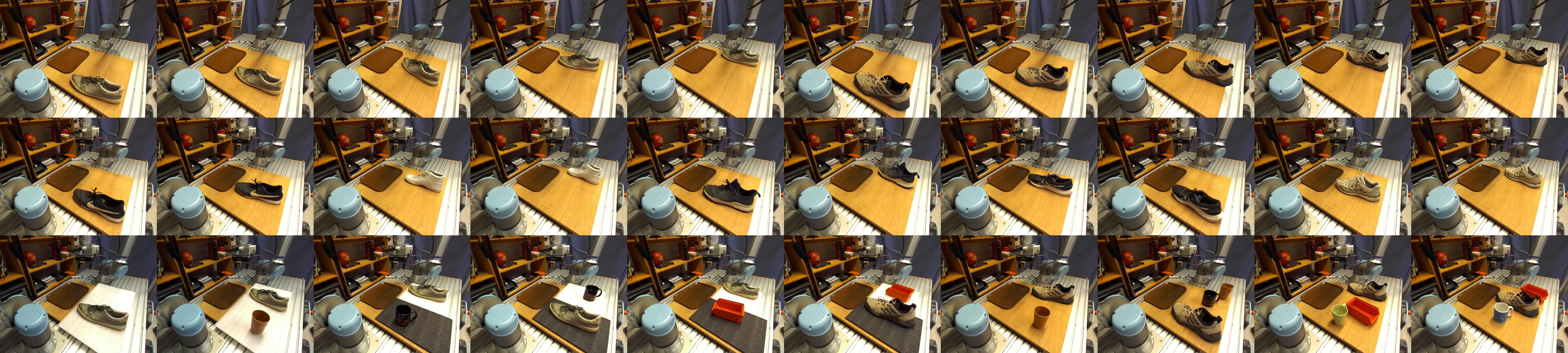}\end{minipage}\\[6pt]
    \begin{minipage}[c]{0.03\linewidth}\centering\rotatebox{90}{Place Mug}\end{minipage}%
    \begin{minipage}[c]{0.97\linewidth}\includegraphics[width=\linewidth]{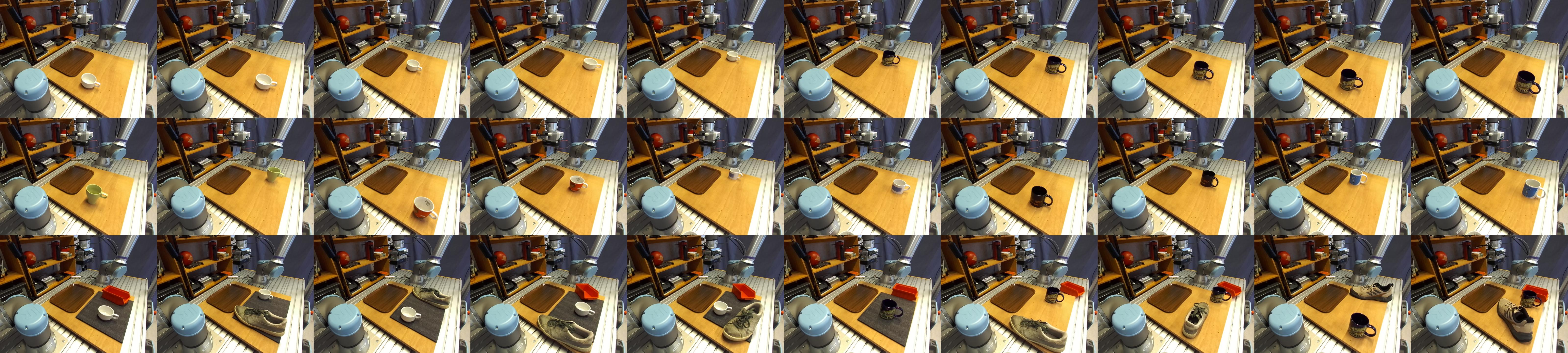}\end{minipage}\\[6pt]
    \begin{minipage}[c]{0.03\linewidth}\centering\rotatebox{90}{Pour Bottle}\end{minipage}%
    \begin{minipage}[c]{0.97\linewidth}\includegraphics[width=\linewidth]{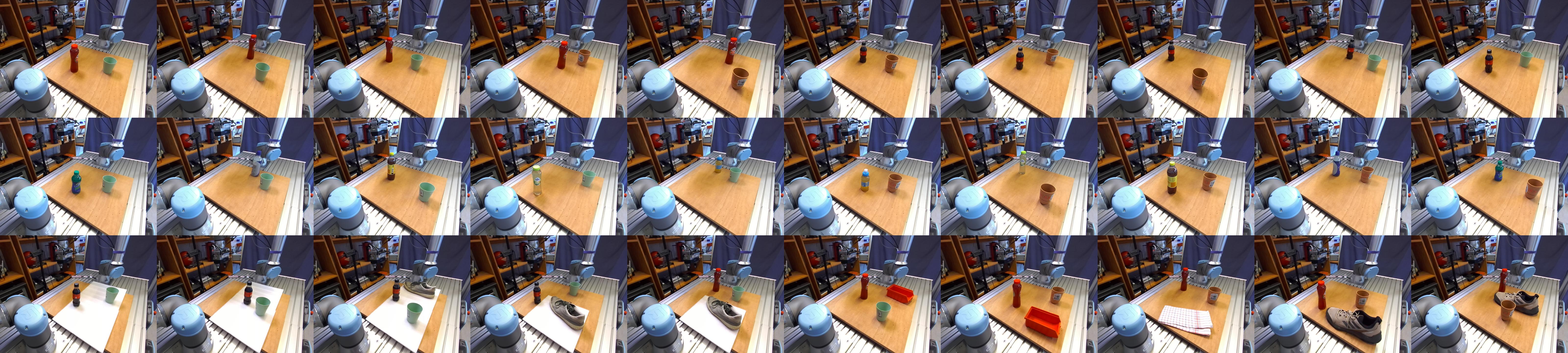}\end{minipage}\\[6pt]
    \begin{minipage}[c]{0.03\linewidth}\centering\rotatebox{90}{Pick Pen}\end{minipage}%
    \begin{minipage}[c]{0.97\linewidth}\includegraphics[width=\linewidth]{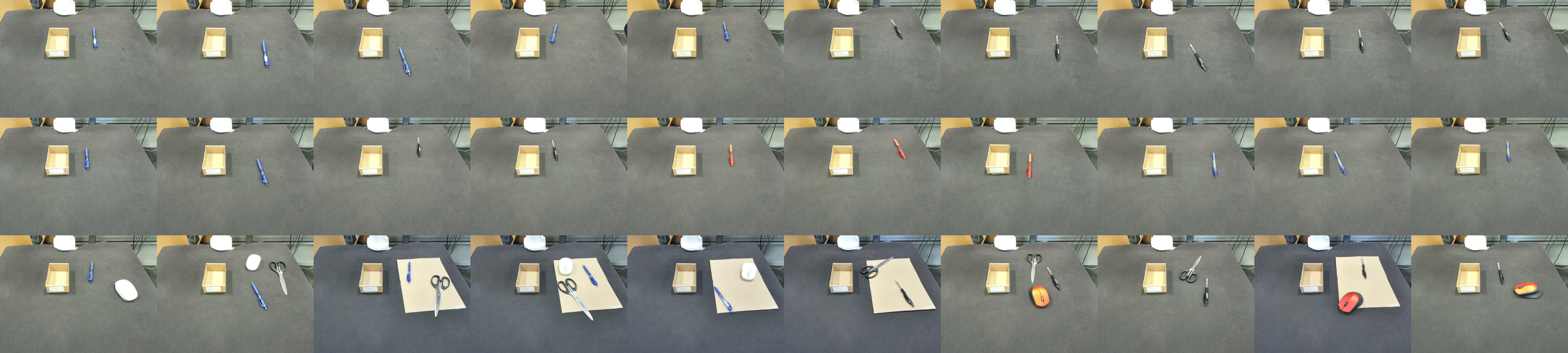}\end{minipage}\\[6pt]
    \begin{minipage}[c]{0.03\linewidth}\centering\rotatebox{90}{Place Mouse}\end{minipage}%
    \begin{minipage}[c]{0.97\linewidth}\includegraphics[width=\linewidth]{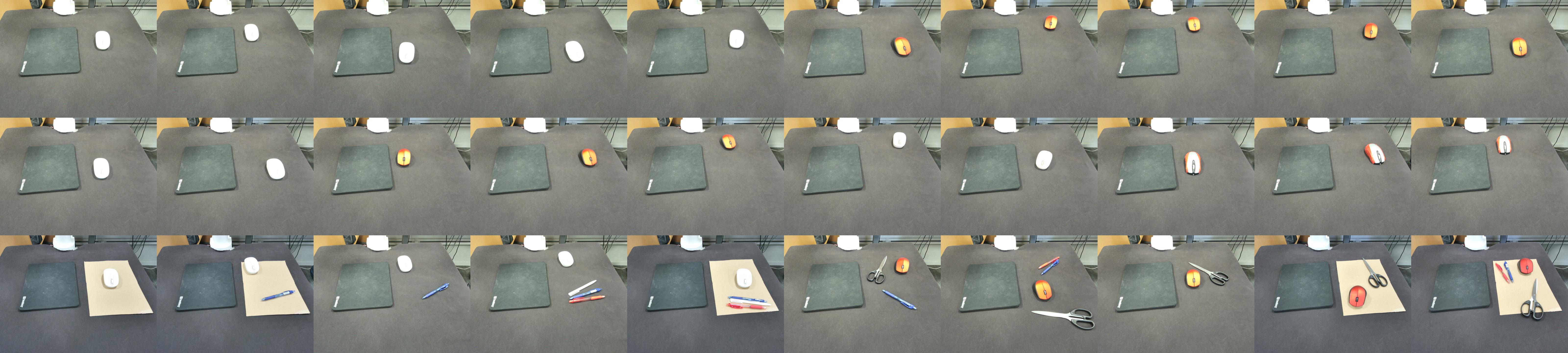}\end{minipage}\\[6pt]
    \caption{Initial scene configurations used for evaluation. For each task, we use 10 in-distribution configurations (first row), 10 unseen object configurations (second row) and 10 scene variation configurations (third row).}
    \label{fig:eval-configurations}
\end{figure*}

\subsection{Unaggregated Results}
\label{sec:appendix-detailed-results}
In this appendix, we provide the unaggregated results for tables in the main text and in Appendix~\ref{sec:appendix-additional-experiments}, broken down by task and evaluation setting (in-distribution, unseen objects, scene variations). We only include tables for which the unaggregated results were not already shown.

\begin{table*}[t]
    \renewcommand{\arraystretch}{1.2}
    \begin{minipage}[t]{0.63\linewidth}
        \centering
        \setlength{\tabcolsep}{4pt}
        \caption{Unaggregated results for Table~\ref{tab:model-comparison}: Comparison of keypoint feature models for the different keypoint extraction methods. The models used in the other experiments in this paper are marked in bold.}
        \label{tab:model-comparison-full}
        \begin{tabularx}{\linewidth}{l *{3}{Y} *{3}{Y} {c}}
            \toprule
            \multirow{2}{*}{\textbf{Method}} & \multicolumn{3}{c}{\textbf{place shoe}} & \multicolumn{3}{c}{\textbf{place mug}} & \multirow{2}{*}{\textbf{Total (/60)}} \\
            \cmidrule(lr){2-4} \cmidrule(lr){5-7}
            & in & obj & scene & in & obj & scene & \\
            \midrule
            RGB~\cite{chi2025diffusion_policy} & 8 & 6 & 3 & 10 & 7 & 0 & 34 \\
            \midrule
            \textbf{RADIOv2.5-B}  & 10 & 6 & 9 & 10 & 8 & 8 & 51 \\
            DIFT$_2$  & 7 & 4 & 2 & 10 & 8& 6 &37  \\
            DINOv3-B & 10 & 4 & 6 & 10 & 5 & 9 & 44  \\
            \midrule
            Instance + RADIOv2.5-B &7 & 6& 6&9 &8 &10 & 46 \\
            \textbf{Instance + DIFT$_2$} & 10 & 6 & 5 & 10 & 7 & 10 & 48 \\
            \midrule
            Tracking + RADIOv2.5-B & 7&6 &7 &9 &6 & 9 & 44 \\
            Tracking + DIFT$_2$ & 9 & 8 & 6 & 10 & 7 & 9 & 49 \\
            \textbf{Tracking + DIFT$_8$} &9 & 6&6 &9 & 9 & 9 & 48 \\
            \bottomrule
        \end{tabularx}
    \end{minipage}%
    \hfill
    \begin{minipage}[t]{0.34\linewidth}
        \centering
        \setlength{\tabcolsep}{5pt}
        \caption{Unaggregated results for Table~\ref{tab:augmentation-comparison}: Impact of Augmentations on KIL.}
        \label{tab:augmentation-comparison-full}
        \begin{tabularx}{\linewidth}{l *{3}{Y} *{3}{Y} {c}}
            \toprule
            \multirow{2}{*}{\textbf{Method}} & \multicolumn{3}{c}{\textbf{pick pen}} & \multicolumn{3}{c}{\textbf{place mouse}} & \multirow{2}{*}{\textbf{Total (/60)}} \\
            \cmidrule(lr){2-4} \cmidrule(lr){5-7}
            & in & obj & scene & in & obj & scene & \\
            \midrule
            None & 7 & 10  & 3 & 1 & 2 & 2 &  25 \\
            Noise  & 6 & 9  & 7 & 10 & 8 & 8 & 48  \\
            ST & 5 & 6  & 4 & 10 & 8 & 9 & 42  \\
            Noise + ST & 7 & 8  & 4 & 10 & 8 & 9 &  46 \\
            \bottomrule
        \end{tabularx}
        \vspace{1em}
        \setlength{\tabcolsep}{2pt}
        \caption{Unaggregated results for Table~\ref{tab:encoder-comparison}: Encoder Comparison.}
        \label{table:encoder-comparison-full}
        \begin{tabularx}{\linewidth}{l *{3}{Y} *{3}{Y} c}
            \toprule
            \multirow{2}{*}{\textbf{Method}} & \multicolumn{3}{c}{\textbf{place shoe}} & \multicolumn{3}{c}{\textbf{place mug}} & \multirow{2}{*}{\textbf{Total (/60)}} \\
            \cmidrule(lr){2-4} \cmidrule(lr){5-7}
            & in & obj & scene & in & obj & scene & \\
            \midrule
            \textbf{ours} & 10 & 6 & 9 & 10 & 8 & 8 & \textbf{51} \\
            None   & 10 & 9 & 9 & 8 & 7 & 8 & \textbf{51} \\
            \hspace{.3em} + sim   & 2 & 0 & 2 & 0 & 0 & 0 & 4 \\
            \hspace{1.2em} - aug & 7 & 6 & 7 & 9 & 5 & 5 & 39 \\
            \bottomrule
        \end{tabularx}
    \end{minipage}
\end{table*}

\begin{table*}[t]
    \renewcommand{\arraystretch}{1.2}
    \begin{minipage}[t]{0.54\linewidth}
        \centering
        \small
        \setlength{\tabcolsep}{5pt}
        \caption{Unaggregated results for Table~\ref{table:2-object-experiments}: Task completion for tasks in which multiple instances of the same object category are handled.}
        \label{table:2-object-experiments-full}
        \begin{tabularx}{\linewidth}{l *{6}{Y} c}
            \toprule
            \multirow{2}{*}{\textbf{Method}} &
            \multicolumn{3}{c}{\textbf{place 2 shoes}} &
            \multicolumn{3}{c}{\textbf{pick 2 pens}} &
            \multirow{2}{*}{\textbf{Total (/60)}} \\
            \cmidrule(lr){2-4} \cmidrule(lr){5-7}
            & in & obj & scene & in & obj & scene & \\
            \midrule
            RGB~\cite{chi2025diffusion_policy}      & 6.5 & 4   & 2.5 & 8.5 & 4.5 & 2.5 & 28.5 \\
            S$^2$~\cite{yang2025S2}       & 4   & 5.5 & 5   & 8   & 7   & 5.5 & 35 \\
            KIL (IM)       & 8.5 & 6   & 7   & 7.5 & 4.5 & 4.5 & 38 \\
            KIL (IN) & 8 & 7.5 & 6 & 7.5 & 5.5 & 2  & 36.5 \\
            KIL (T) & 7.5 & 4.5 & 3.5 & 6 & 5.5 & 4 & 31 \\
            \bottomrule
        \end{tabularx}
    \end{minipage}%
    \hfill
    \begin{minipage}[t]{0.43\linewidth}
        \centering
        \setlength{\tabcolsep}{5pt}
        \caption{Unaggregated results for Table~\ref{tab:more-rotations-experiment}: Illustration of performance drop when increasing the range of initial orientations for task objects.}
        \label{tab:more-rotations-experiment-full}
        \begin{tabularx}{\linewidth}{l*{3}{Y}*{3}{Y}c}
            \toprule
            \multirow{2}{*}{\textbf{Method}} & \multicolumn{3}{c}{\textbf{pick pen ($\pm 180^\circ$)}} & \multicolumn{3}{c}{\textbf{place mouse ($\pm 180^\circ$)}} & \multirow{2}{*}{\makecell{\textbf{Total}\\\textbf{(/60)}}} \\
            \cmidrule(lr){2-4} \cmidrule(lr){5-7}
            & in & obj & scene & in & obj & scene & \\
            \midrule
            RGB & 6 & 7  & 0 & 8 & 7 & 0 &  28 \\
            S$^2$  & 7 & 6  & 5 & 9 & 10 & 5 & \textbf{42} \\
            KIL (IM) & 6 & 4  & 2 & 5 & 3 & 1 & 21 \\
            KIL (T) & 6 & 5  & 4 & 5 & 3 & 6 &  29 \\
            \bottomrule
        \end{tabularx}
    \end{minipage}
\end{table*}

\fi 

\end{document}